\documentclass[
a4paper,
]{ipi}
\usepackage{mathtools}
\usepackage{amsmath}
\usepackage{todonotes}
  \usepackage{paralist}
  \usepackage{graphics} 
  \usepackage{epsfig} 
\usepackage{graphicx}  \usepackage{epstopdf}

\usepackage{caption}
\usepackage{subcaption}
\usepackage{makecell}
\usepackage{csquotes}
\usepackage[export]{adjustbox}
\usepackage{float}

 \usepackage[colorlinks=true]{hyperref}
\hypersetup{urlcolor=blue, citecolor=red}

\def\currentvolume{X}

\theoremstyle{definition}

\title[Extreme Deblurring of Text Images]{Let's Enhance: A Deep Learning Approach to Extreme Deblurring of Text Images}

\author[Trippe, Genzel, Macdonald, and März]{}


 \email{theophil.trippe@hotmail.com}
 \email{martingenzel@gmail.com}
 \email{macdonald@math.tu-berlin.de}
 \email{maxmaerz@amazon.de}

\begin{document}
\maketitle

\centerline{\scshape Theophil Trippe}
\medskip
{\footnotesize
   \centerline{Technische Universität Berlin, Department of Mathematics, Berlin, Germany}
   \centerline{(Corresponding author)}
} 

\medskip

\centerline{\scshape Martin Genzel}
\medskip
{\footnotesize
 \centerline{Helmholtz-Zentrum Berlin für Materialien und Energie\footnote{This work was done while at Utrecht University.},}
 \centerline{Department of Optics and Beamlines, Berlin, Germany}
}

\medskip

\centerline{\scshape Jan Macdonald}
\medskip
{\footnotesize
   \centerline{Technische Universität Berlin, Department of Mathematics, Berlin, Germany}
}

\medskip

\centerline{\scshape Maximilian März}
\medskip
{\footnotesize
\centerline{Amazon Web Services\footnote{This work was done prior to joining Amazon.}, Berlin, Germany}
}

\bigskip


\begin{abstract}
This work presents a novel deep-learning-based pipeline for the inverse problem of image deblurring, leveraging augmentation and pre-training with synthetic data.
Our results build on our winning submission to the recent \emph{Helsinki Deblur Challenge 2021}, whose goal was to explore the limits of state-of-the-art deblurring algorithms in a real-world data setting.
The task of the challenge was to deblur out-of-focus images of random text, thereby in a downstream task, maximizing an optical-character-recognition-based score function.
A key step of our solution is the data-driven estimation of the physical forward model describing the blur process. 
This enables a stream of synthetic data, generating pairs of ground-truth and blurry images on-the-fly, which is used for an extensive augmentation of the small amount of challenge data provided.
The actual deblurring pipeline consists of an approximate inversion of the radial lens distortion (determined by the estimated forward model) and a \mbox{U-Net} architecture, which is trained end-to-end.
Our algorithm was the only one passing the hardest challenge level, achieving over $70\%$ character recognition accuracy.
Our findings are well in line with the paradigm of data-centric machine learning, and we demonstrate its effectiveness in the context of inverse problems.
Apart from a detailed presentation of our methodology, we also analyze the importance of several design choices in a series of ablation studies.
The code of our challenge submission is available under \url{https://github.com/theophil-trippe/HDC_TUBerlin_version_1}.

\medskip

\textit{This article has been published in a revised form in ``Inverse Problems and Imaging'' (DOI: \url{https://doi.org/10.3934/ipi.2023019}). This version is free to download for private research  and study only. Not for redistribution, re-sale or use in derivative works.}
\end{abstract}

\section{Introduction}

\emph{Image blurring} is an undesirable distortion effect that occurs in many image acquisition processes.
Possible causes are a lack of camera focus~\cite{defocus_blur}, rapid object movement~\cite{motion}, light scattering effects~\cite{scatter}, or unstable camera operations.
The same effect is, for instance, also observed in near- and farsightedness, where the retina of the human eye loses the ability to focus.
The aim of \emph{deblurring} is the inversion of this effect, i.e., the retrieval of a sharp image.
Deblurring has numerous practical applications, such as computed tomography~\cite{CT}, consumer photography~\cite{consumer_photo}, or radar imaging~\cite{radar}.
In general, any optical imaging process is susceptible to blurring artifacts and can benefit from deblurring techniques.

Mathematically, deblurring is an important example of an \emph{inverse problem} (see Section~\ref{sec:problem} for more details).
In its most simple form, the forward model of a blurring process is a linear convolution with a spatially localized blur kernel.
Already in this case, the inverse problem can be highly \emph{ill-posed}, depending on the size of the blur kernel and the decay of singular values of the associated convolution operator.
Even more challenging is the setting of \emph{blind deconvolution} with an unknown blur kernel~\cite{hansen,levin09}.

\begin{figure} 
	\centering
	\begin{tabular}{c@{\,}c@{\,}c@{\,}c}
		\textbf{\ level \ } & blurry image & \parbox{2.7cm}{\centering our \\ reconstruction\\[.5em] } & \parbox{2.7cm}{\centering sharp image \\ (ground truth)\\[.5em] } \\
		\vspace{0.15em}
		
		\textbf{4} & 
		\includegraphics[valign=c,width=2.7cm]{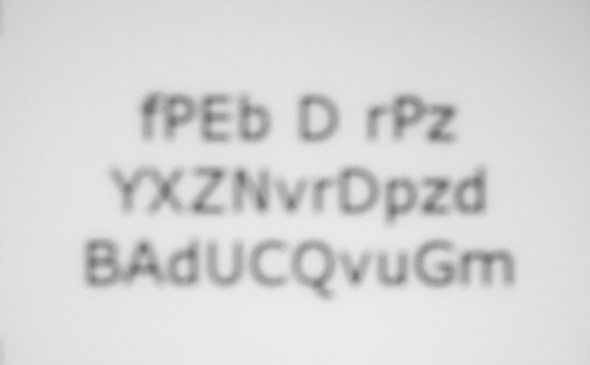} & 
		\includegraphics[valign=c,width=2.7cm]{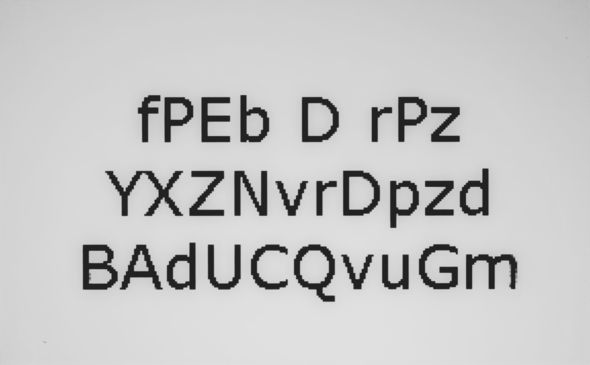} &
		\includegraphics[valign=c,width=2.7cm]{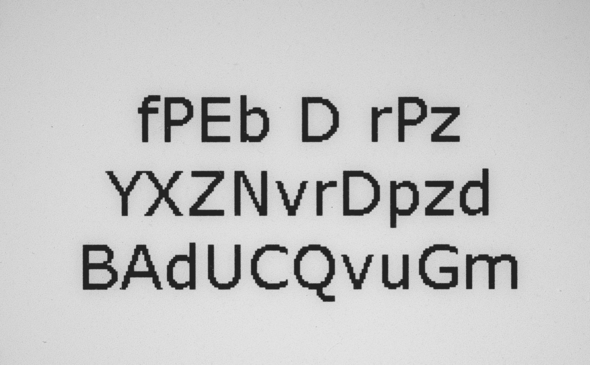} \\ 
		\vspace{0.15em}
		
		\textbf{9} & 
		\includegraphics[valign=c,width=2.7cm]{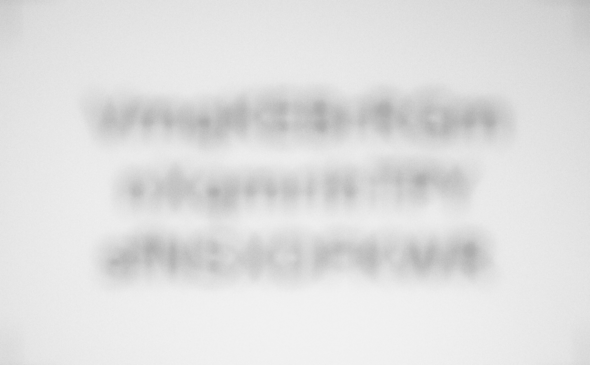} & 
		\includegraphics[valign=c,width=2.7cm]{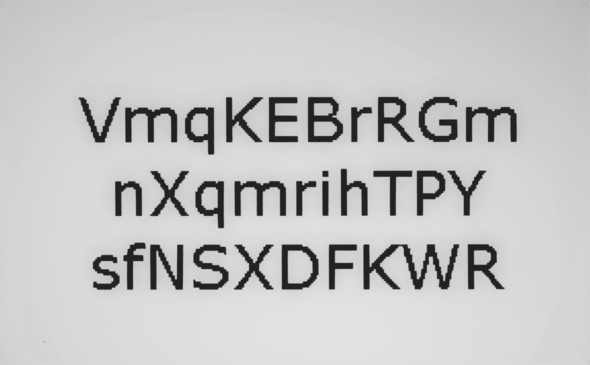} &
		\includegraphics[valign=c,width=2.7cm]{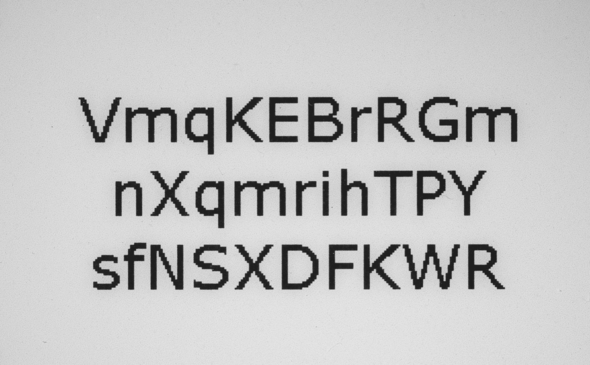} \\ 
		\vspace{0.15em}
		
		\textbf{14} & 
		\includegraphics[valign=c,width=2.7cm]{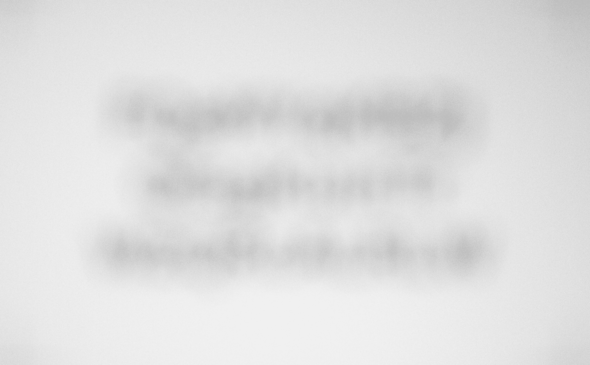} & 
		\includegraphics[valign=c,width=2.7cm]{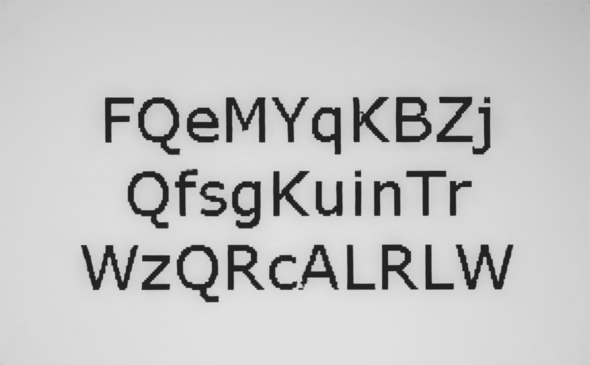} &
		\includegraphics[valign=c,width=2.7cm]{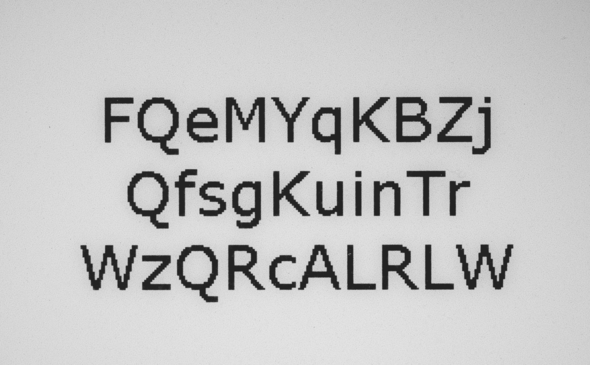} \\      
		\vspace{0.15em}
		
		\textbf{19} & 
		\includegraphics[valign=c,width=2.7cm]{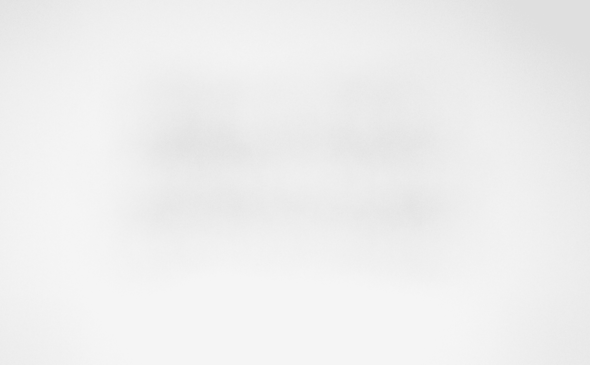} & 
		\includegraphics[valign=c,width=2.7cm]{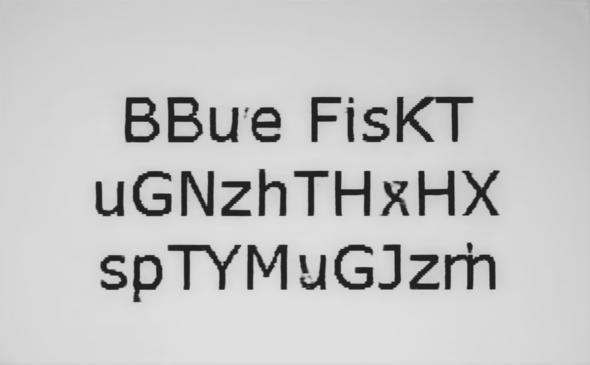} &
		\includegraphics[valign=c,width=2.7cm]{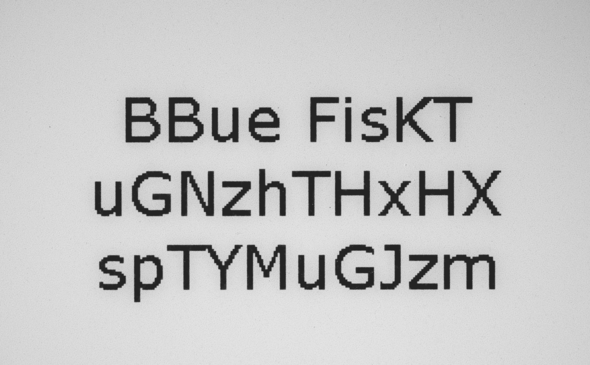}
	\end{tabular}
	\caption{Examples from the HDC dataset with different blur levels (left and right column). The severity of blurring increases with each level ($20$ in total, ranging from $0$ to $19$). The center column shows the corresponding reconstructions with our deep-learning-based pipeline (examples taken from the validation set).}
	\label{fig:glimpse}
\end{figure}

To make further progress in this research field, the \emph{Helsinki Deblur Challenge 2021} (\emph{HDC}) was initiated to explore the limits of state-of-the-art data-driven deblurring methods in a blind deconvolution setting \cite{HDC_website}.
For this purpose, the organizers of the HDC have set up $20$ blur kernels corresponding to real-world photographic measurements with an increasing amount of misfocus, thereby making the inverse problem more and more ill-posed, up to extreme cases.
The task was to deblur images of (random) text, so that a subsequent optical character recognition (OCR) becomes as accurate as possible.
Fig.~\ref{fig:glimpse} gives a first impression of the HDC task and our results; see Section~\ref{sec:challenge} for more details on the challenge setup.

Not explicitly knowing the forward model and the availability of relatively little ground-truth data lead to a very challenging problem.
Especially the clever design of the HDC measurement setup, which allowed the acquisition of well-aligned pairs of real-world blurry and sharp images, sparked our interest in participating.
The present article is devoted to our winning submission to the HDC.
Besides a description of the underlying methodology, our main objective is to distill several key insights regarding the design and training of our deep learning approach, which is of broader interest.

\subsection{Our Approach in a Nutshell}

\begin{figure}
	\centering
	\includegraphics[width=.9\linewidth]{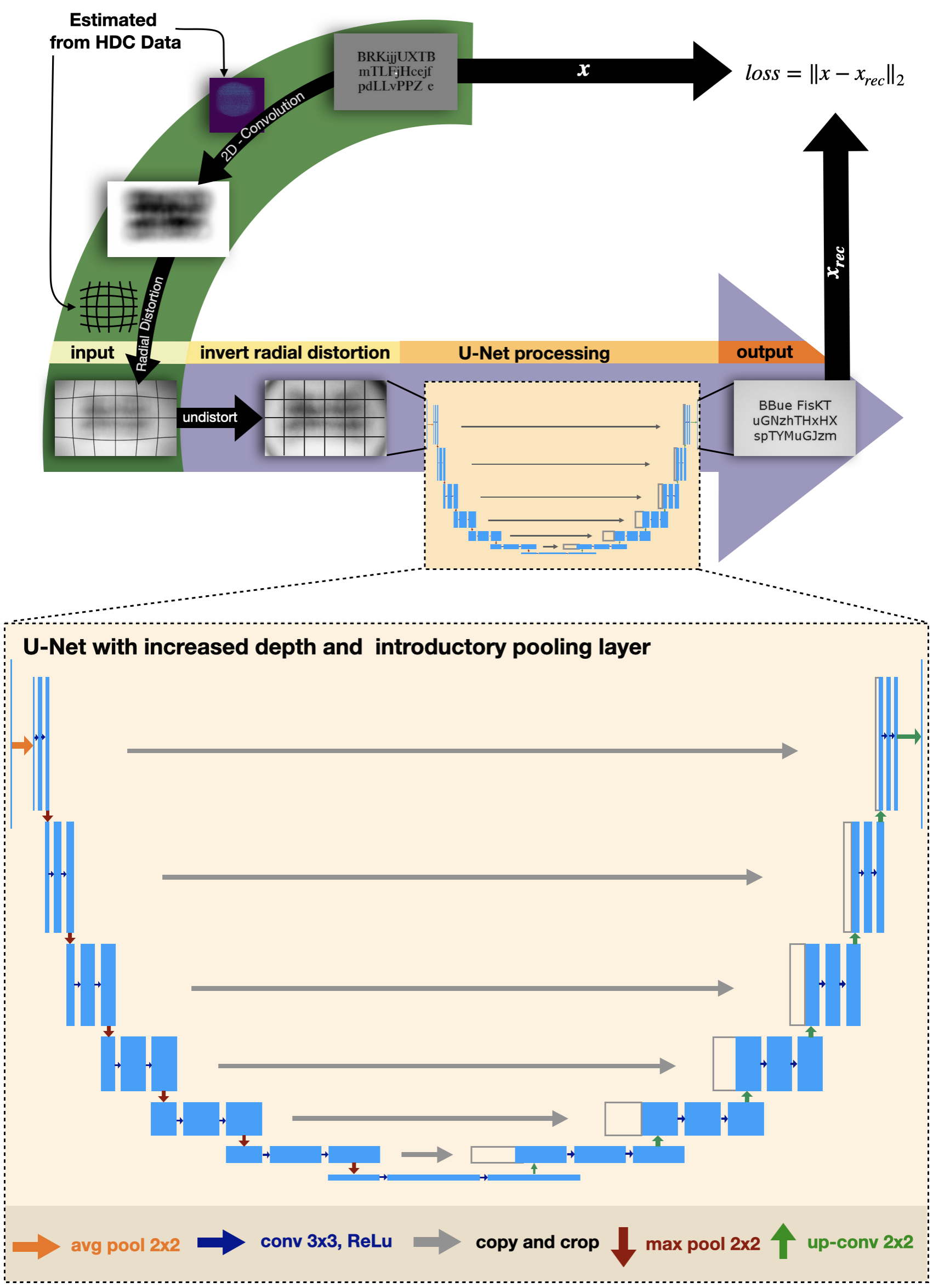}  
	\caption{Schematic depiction of our deblurring pipeline. Top left (green): Data is synthesized for augmenting the available HDC training data (the details of the background removal step described in Section~\ref{subsec:fwd} are omitted here; see Fig.~\ref{fig:background} for a precise depiction of that step).
		Center (purple): The reconstruction pipeline consists of an inverse lens distortion and a modified U-Net architecture. Bottom: The modified architecture differs from the vanilla U-Net (cf.~\cite{U-Net}) by adding an introductory pooling layer and additional down- and up-sampling levels, to increase the overall receptive field.}
	\label{fig:pipeline}
\end{figure}

Over the last few years, \emph{deep convolutional neural networks} have been successfully used to solve ill-posed inverse problems in imaging applications~\cite{data_driv_ip, DL4IP}.
Training these networks typically requires large amounts of data pairs consisting of ground-truth signals (sharp images in HDC) and their corresponding measurements (blurry images in HDC).
With only 200 samples per blur level, the training data provided by the HDC is rather limited in that respect.
Still aiming at a \emph{deep-learning-based approach}, our key idea is to infer the unknown blur process from the data and to use it for augmenting the dataset with additional \emph{synthetic} image pairs; see the green region in Fig.~\ref{fig:pipeline}.
We therefore start with the modeling and estimation of a forward operator that accurately reflects the physical reality of the blur process.
Our basic model consists of a single blur kernel that is rendered spatially variant by a radial lens distortion (cf.~Section~\ref{sec:problem}). 

The actual deblurring pipeline is schematically depicted in the purple region of Fig.~\ref{fig:pipeline}. It is based on an approximate inversion of the radial lens distortion (not learnable) followed by a slightly modified version of the popular U-Net architecture \cite{U-Net}.
The neural network training is done \emph{end-to-end} (i.e., all learnable parameters are optimized simultaneously) and carried out in two phases: first, an extensive \emph{pre-training} on the simulated synthetic data, and second, a \emph{fine-tuning} on the provided real-world challenge data.

Beyond winning the challenge, our work emphasizes the paradigm of \emph{data-centric} machine learning \cite{ng-datacentric} as we use a very common neural network architecture and rather focus on generating suitable training data.
Here, a problem-specific difficulty is the estimation of the forward model, which does not only allow us to create a data stream for pre-training but also to decompose the deblur process into two steps: Resolving the lens distortion first and performing the deconvolution afterwards.
Our main contributions can be summarized as follows:

\begin{enumerate}[(i)]
	\item We explore the limits of deblurring by demonstrating that severely blurred images can be recovered with an end-to-end deep learning pipeline. The accuracy of a subsequent OCR confirms that substantial visual information is reconstructed with this procedure (see also Fig.~\ref{fig:glimpse}).
	While the effectiveness of deep learning in deblurring is well known, to the best of our knowledge, its success in such extreme situations has not been reported yet.
	\item We strengthen the paradigm of data-centric machine learning, finding that simple building blocks (end-to-end learning, U-Net, etc.) in conjunction with careful data augmentation and pre-training can lead to superior results.
	This stands in contrast to the model-centric viewpoint of machine learning, which typically requires more architecture engineering and expert knowledge on the underlying problem.
	\item We provide unambiguous evidence that state-of-the-art results for inverse problems with real-world data can be achieved by training a neural network based on mostly synthetic data. However, an important prerequisite for this is an accurate estimate of the forward model.
\end{enumerate}

\subsection{Outline}

The remainder of this article is organized as follows. Section~\ref{sec:back} formally introduces the deblurring problem, briefly discusses relevant related literature, and gives an overview of the HDC setup.
Section~\ref{sec:meth} provides a conceptual description of our methodology and learning pipeline.
Our main results are then presented in Section~\ref{sec:res} as well as several accompanying ablation studies.
We conclude with a short discussion in Section~\ref{sec:disc}.

\section{Background and Problem Definition}\label{sec:back}
This section introduces two variants of the deblurring problem more formally (Section~\ref{sec:problem}). We then discuss related works (Section~\ref{sec:literature}) and briefly present the specific challenge setup and rules (Section~\ref{sec:challenge}).

\subsection{Formalizing the Deblurring Problem}\label{sec:problem}
We start by introducing a continuous formulation of spatially invariant blurring and deblurring (Section~\ref{sec:problem:invar}). This can be generalized by allowing spatially varying blur kernels, e.g., caused by radial lens distortion (Section~\ref{sec:spatial_variance}).
Finally, we discuss some aspects of discretization (Section~\ref{sec:problem:discrete}).

\subsubsection{Spatially Invariant Continuous Blurring}\label{sec:problem:invar}
Let $I_s \in L^2([0,1] \times [0,1])$ be a continuous (sharp) image. In its most generic form, a \emph{blurring} of $I_s$ is represented as a bounded linear operator $B$ on $L^2([0,1] \times [0,1])$ that acts in a ``spatially local'' manner.
In the simplest case, $B$ amounts to a spatially invariant convolution
\begin{equation}\label{eq:blur_invariant}
	I_b = B(I_s) = I_s * k_B
\end{equation}
where $I_b \in L^2([0,1] \times [0,1])$ denotes the blurred image and $k_B \in L^2([0,1] \times [0,1])$
is the \emph{blur kernel} associated with $B$. More specifically, the blurred image at location $(x,y)\in[0,1] \times [0,1]$ is given by
\begin{equation}\label{eq:blur_xy_invariant}
	I_b(x,y) = \iint I_s(\tau, \theta) k_B(x-\tau, y - \theta) \, \mathrm{d}\tau \, \mathrm{d}\theta
\end{equation}
where some appropriate boundary conditions and corresponding integration domains are assumed.

We consider the inverse problem of recovering the sharp image $I_s$ from the (possibly noisy) measurements $I_b + e$, where $e \in L^2([0,1] \times [0,1])$ denotes additive noise.
In general, this is an ill-posed problem, and classical reconstruction methods include variational approaches such as Tikhonov regularization~\cite{tikhonov} and total-variation minimization~\cite{rof92}; see Section~\ref{sec:literature}.
In the case of an unknown blur kernel $k_B$, the problem is referred to as \emph{blind deblurring} or \emph{blind deconvolution}.

\subsubsection{Spatially Variant Blurring and Radial Distortion}\label{sec:spatial_variance}
\begin{figure}
	\centering
	\includegraphics[width=.9\linewidth]{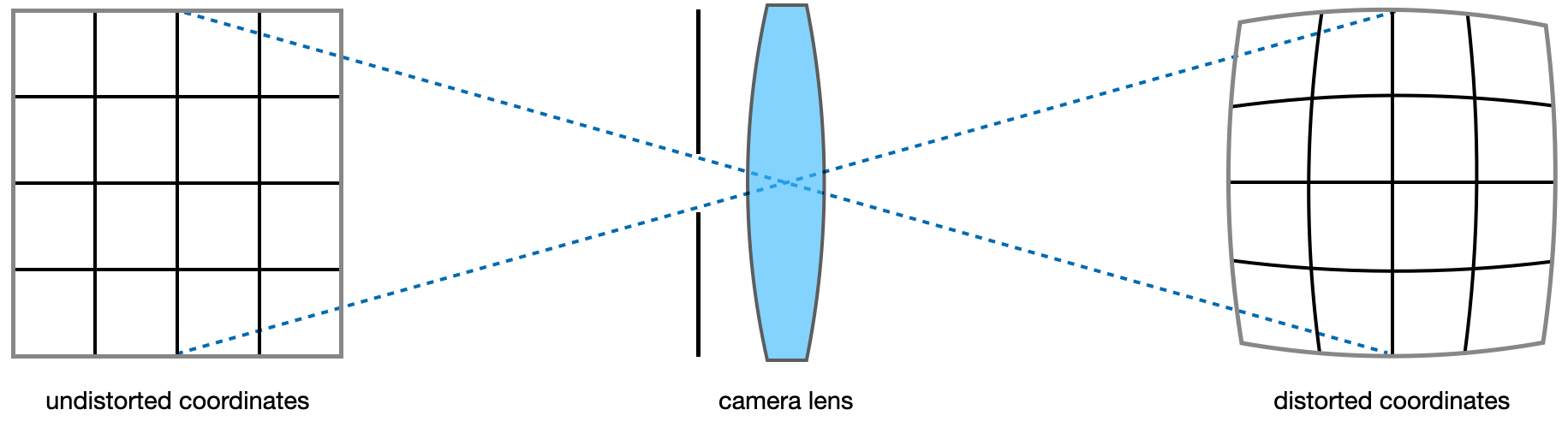}
	\caption{Visualization of the coordinate deformation caused by radial lens distortion.}
	\label{fig:convex_lens}
\end{figure}

In many applications, the simple convolutional model \eqref{eq:blur_invariant} is not sufficient to accurately describe the physical blur process.
Rather than being determined by a single blur kernel, \emph{spatially variant blur} is described by a blur kernel function $k_B \colon [0,1] \times [0,1] \rightarrow L^2([0,1] \times [0,1])$, which determines the blur kernel at all positions $(x,y) \in [0,1] \times [0,1]$. In this case, \eqref{eq:blur_xy_invariant} becomes
\begin{equation}\label{eq:spat_var}
	I_b(x, y) = \iint I_s(\tau, \theta) k_B(x,y)(x-\tau, y - \theta) \, \mathrm{d}\tau \, \mathrm{d}\theta.
\end{equation}
While the spatial variation in the formal description above can be arbitrary, in applications, $k_B$ often exhibits regularity properties such as partial smoothness, low-dimensional approximation, or symmetry.
Understanding the structure of these regularities can be crucial to account for the spatial variance in the reconstruction.
One important example that is relevant to our deblurring approach is spatial variance caused by \emph{radial lens distortion}.
It naturally occurs due to the convex shape of camera lenses~\cite{demtroeder} and can intuitively be understood as a concatenation of a spatially invariant convolution with some (fixed) kernel $k_B\in L^2([0,1] \times[0,1])$ and a continuous deformation $d \colon \mathbb{R}^2 \rightarrow \mathbb{R}^2$ of the underlying $(x,y)$-coordinates.
The deformation is radially symmetric with respect to some center $(x_0, y_0) \in \mathbb{R}^2$, see Fig.~\ref{fig:convex_lens} for a visualization.
A simplified\footnote{The full distortion model of Brown (also known as Brown-Conrady model) also accounts for tangential distortion effects (called decentering distortion) in addition to the purely radial distortion.} variant of \emph{Brown's even-order polynomial model} for the radial distortion \cite{brown1966} is based on a Taylor expansion around $(x_0,y_0)$ and describes the relation between coordinates in the distorted and undistorted image through the transformation
\begin{align}\label{eq:distort}
	d(x,y) = (x_0,y_0) + \left(1+\sum_{q=1}^{Q} K_q \|(x,y)-(x_0,y_0)\|_2^{2q}\right) \cdot (x-x_0,y-y_0),
\end{align}
where $K_q\in\mathbb{R}$ are the radial distortion coefficients and $Q\in\mathbb{N}$ is the maximal order of the expansion.
We therefore obtain the image intensity $I_b(x, y)$ of the distorted blurry image $I_b$ at $(x,y)$ as the intensity of the ideal blurry image $(I_s * k_B)$ (without distortion) at the corresponding distorted coordinates $d(x,y) = (\tilde{x}, \tilde{y})$. Hence, we can write this special case of a spatially variant blur model \eqref{eq:spat_var} as
\begin{equation}\label{eq:cont_model_full}
	I_b(x, y) = (I_s * k_B)(d(x,y)) = \iint I_s(\tau, \theta) k_B(\tilde{x}-\tau, \tilde{y} - \theta) \, \mathrm{d}\tau \, \mathrm{d}\theta.
\end{equation}
For small distances to the center $(x_0,y_0)$, the distortion $d$ can be approximately inverted (``undistorted'') by the transformation
\begin{equation}\label{eq:undistort}
	u\colon\mathbb{R}^2\to\mathbb{R}^2, \ (\tilde{x},\tilde{y}) \mapsto (x_0,y_0) + \frac{(\tilde{x}-x_0,\tilde{y}-y_0)}{1+\sum_{q=1}^{Q} K_q \|(\tilde{x},\tilde{y})-(x_0,y_0)\|_2^{2q}},
\end{equation}
with the same coefficients $K_q$ as before. This corresponds to a higher order variant of the \emph{division model} for distortion correction as described in~\cite{division_distortion}. However, note that the distance $\|(\cdot,\cdot)-(x_0,y_0)\|_2$ to the center $(x_0,y_0)$ is measured with respect to $(x,y)$-coordinates in \eqref{eq:distort}, while it is measured with respect to $(\tilde{x},\tilde{y})$-coordinates in \eqref{eq:undistort}. Hence, $u\neq d^{-1}$ is not an exact inversion of $d$ and a distortion correction via $u$ will degrade further away from the center $(x_0,y_0)$. This effect becomes more severe for large distortion coefficients. Nevertheless, we will make use of this approximate inversion in Section~\ref{sec:step_3}.

\subsubsection{Discrete Deblurring Problems}\label{sec:problem:discrete}
While the continuous version of the deblurring problem is elegant and allows us to concisely describe distortion effects, applications usually work with discrete images, which is the setting we will consider from now on. In the discrete case, the sharp and blurry images are given by matrices $\mathbf{x}, \mathbf{y} \in \mathbb{R}^{m \times n}$ where $m, n \in \mathbb{N}$ denote the number of pixels in the vertical and horizontal spatial dimension, respectively.\footnote{We will usually not distinguish between image matrices in $\mathbb{R}^{m\times n}$ and their vectorized forms in $\mathbb{R}^{mn}$. It will be clear from the context whether discrete images are viewed as matrices or vectors.} A simple convolution can then be described by a discrete blur kernel $\mathbf{k}_B \in \mathbb{R}^{s \times s}$ ($s \in 2\mathbb{N} + 1$) through
\begin{equation}\label{eq:disc_conv}
	\mathbf{y}_{i,j} = (\mathbf{k}_B * \mathbf{x})_{i,j} = \sum_{h, \ell=-\left\lfloor\frac{s}{2}\right\rfloor}^{\left\lfloor\frac{s}{2}\right\rfloor} {(\mathbf{k}_B)}_{\left\lfloor\frac{s}{2}\right\rfloor+h,\left\lfloor\frac{s}{2}\right\rfloor+\ell} \cdot \mathbf{x}_{i+h, j+\ell} 
\end{equation}
where appropriate boundary conditions on $\mathbf{x}$ are assumed such that $\mathbf{x} \in \mathbb{R}^{m \times n}$ is embedded into $\mathbf{x} \in \mathbb{R}^{\mathbb{Z} \times \mathbb{Z}}$.
For commonly used boundary conditions, such as zero padding, periodic, or reflexive continuation, the corresponding operator matrix $\mathbf{B} \in \mathbb{R}^{mn \times mn}$ acting on the vectorized images is highly structured and exhibits favorable diagonalization properties. In the case of periodic boundary conditions, for instance, $\mathbf{B}$ is a block circulant matrix with circulant blocks (BCCB), whose spectral decomposition can be efficiently computed via the Fast Fourier Transform~\cite{hansen}.

The radial distortion model $d$ in \eqref{eq:distort} is inherently continuous as it involves the continuous distortion of coordinates. Our discretized implementation is therefore based on interpolating between the discrete pixels values, applying the continuous radial distortion $d$ (or undistortion $u$) on these interpolated values, and then sampling the result back to the discrete $m \times n$ grid to obtain the discrete distorted (undistorted) image.

Combining the discrete convolution \eqref{eq:disc_conv} with a discretized version of the radial lens distortion results in a discretization of the overall blur model \eqref{eq:cont_model_full}.

\subsection{Related works}\label{sec:literature}
As mentioned above, discrete convolutions with common boundary conditions have favorable diagonalization properties~\cite{hansen} that can be exploited for efficient implementations of traditional regularization techniques.
A prototypical regularization strategy involves solving a \emph{variational problem} of the form
\begin{equation*}
	{\arg \min}_\mathbf{x} \, \tfrac{1}{2} \|\mathbf{B} \mathbf{x}-\mathbf{y}\|_2^2+ \lambda R(\mathbf{x})
\end{equation*}
where $R : \mathbb{R}^{m\times n} \to \mathbb{R}$ is a (convex) penalty function and $\lambda \geq 0$ is a tunable hyperparameter.
A classical choice for $R$ is a Tikhonov regularizer $R(\mathbf{x}) = \|\Gamma \mathbf{x}\|_2^2$ \cite{tikhonov}, which improves the numerical conditioning of the least squares problem while simultaneously allowing for a ``smoothness'' prior through the operator~$\Gamma$.
As (sharp) images are often governed by edge-like structures, another very effective regularization approach is to encourage images with a \emph{sparse} gradient.
This can be achieved by a \emph{total-variation} functional of the form $R(\mathbf{x}) = \|W \mathbf{x}\|_1$ where $W$ is a suitable finite-difference operator \cite{rof92, tv_based}.

Several participating teams of the HDC have indeed made use of such classical variational approaches to address the inverse problem, e.g., see \cite{git_HDC_TV, git_HDC_prox, git_HDC_TV_2}.
However, these methods are limited in blind-deblurring scenarios as they require explicit knowledge of the forward operator $\mathbf{B}$.
When estimating the forward operator, many participating teams have assumed a spatially invariant convolution, which usually only works well under mild distortions and negligible spatial variance. 
For a more extensive introduction to classical image deblurring methods, we refer to the textbook~\cite{hansen}.

A more recent line of research on inverse problems builds on data-driven algorithms, especially \emph{deep learning} techniques.
Arguably, convolutional neural networks (CNNs) are the most widely used type of models for imaging applications, and related inverse problems are no exception in that respect; see \cite{data_driv_ip,DL4IP} for comprehensive overviews.
For the specific problem of image deblurring, Zhang et al.~provide an extensive survey on existing methods~\cite{deepl_debl}.
Their taxonomy of deep learning approaches is roughly summarized in the following list:
\begin{enumerate}
	\item \emph{Deep autoencoders (DAE)} learn to extract relevant features of an image by the \emph{encoder} part of the network and then to reconstruct it from this representation with the \emph{decoder} part of the network \cite{Kingma:2013aa}.
	This is based on the idea that the essential information of natural (sharp) images can be compressed to very few feature variables.
	When adapted to deblurring problems, the encoder takes a blurry image as input, while the decoder is supposed to reconstruct its sharp counterpart.
	A common choice of architecture for this approach is the U-Net and variants thereof, e.g., see~\cite{DAE_1, DAE_2, DAE_3}, whose skip connections allow for an information transfer between the encoder and decoder at different resolution levels (see Fig.~\ref{fig:pipeline}).
	\item \emph{Generative adversarial networks (GANs)} are generative models, whose training is inspired by a zero-sum game between a generator and a discriminator neural network \cite{Goodfellow_2020}.
	In the context of deblurring, the generator is supposed to synthesize sharp images that are indistinguishable from real sharp images for the discriminator~\cite{GAN_3, GAN_1}.
	The generator of a GAN is then typically used as a data-driven prior in a least squares reconstruction.
	\item \emph{Cascaded networks} consist of a sequence of concatenated (simple) networks as building blocks, thereby forming a much deeper architecture. These cascades are inspired by \emph{unrolling} classical iterative regularization methods and often incorporate the forward operator in their sub-blocks~\cite{cascaded_1,amj18,ao18,kl10}.
	Cascaded networks are typically trained end-to-end, where the sub-blocks may share their weights to reduce memory requirements. 
	Alternatively, the sub-blocks can be trained individually to circumvent limitations arising in the training process~\cite{cascaded_2}.
	\item \emph{Multi-scale networks} attempt to recover low-resolution features first and then progressively move towards higher image resolutions. 
	More specifically, a coarse-to-fine scheme may start with generating a sharp image at $1/2^{h}$ of the intended resolution for some $h \in \mathbb{N}$ and then incrementally increases the resolution to $(1/2^{h-i})_{i=1}^h$, where the previously computed (low resolution) reconstructions are used as inputs, e.g., see~\cite{GAN_2, git_HDC_multi_scale}.
	\item \emph{Reblurring networks} involve a blurring network, which is trained to generate blurry images from sharp ones, thereby mimicking the (potentially unknown) blur operator. This is followed by a deblurring network which tries to invert this process.
	Both networks can be trained jointly in an end-to-end fashion, e.g., see~\cite{reblu_1, reblu_2}.
\end{enumerate}

Conceptually, the last approach is most similar to ours.
However, an important difference is that we use a model-based parametrization of the forward operator instead of a generic network, while the blurring only affects the data generation part of our pipeline.
Our actual reconstruction network (undistortion and a slightly modified \mbox{U-Net}) is related to the above DAE approach.
Therefore, a key finding of this paper is that even relatively simple architectures can produce state-of-the-art results if they are trained on carefully designed datasets or streams. 
In particular, our network does not explicitly incorporate the (estimated) blur kernel, as it would be the case in more advanced schemes like cascaded networks.
We refer to Section~\ref{sec:meth} for a detailed presentation of our methodology.

\subsection{Challenge Setup}\label{sec:challenge}
We now describe the specific setup of the HDC in more detail. The challenge was initiated and organized by the \emph{Finnish Invere Problem Society} (FIPS)~\cite{HDC_website}, calling for state-of-the-art algorithms to deblur images of text. It took place from July 31 to September 30, 2021. In this time frame, 17 international teams have participated and 15 methods were submitted for a final evaluation.

\begin{figure}
	\centering
	\includegraphics[width=.5\linewidth]{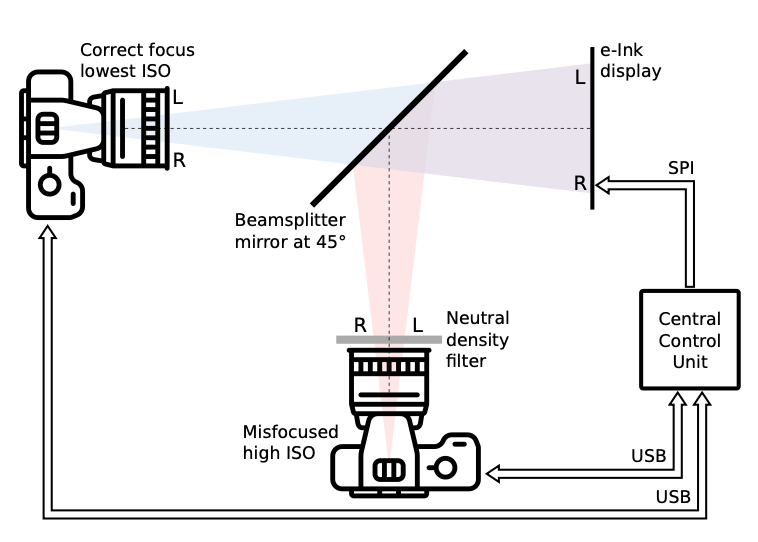}  
	\caption{Illustration of the HDC experimental setup taken from~\cite{HDC_Data}. Two identical photo cameras target the same e-ink display with the help of a half-transparent beamsplitter mirror.}
	\label{fig:photo_setup}
\end{figure} 

\subsubsection{Dataset}
\begin{figure}
	\centering
	\includegraphics[width=.7\linewidth]{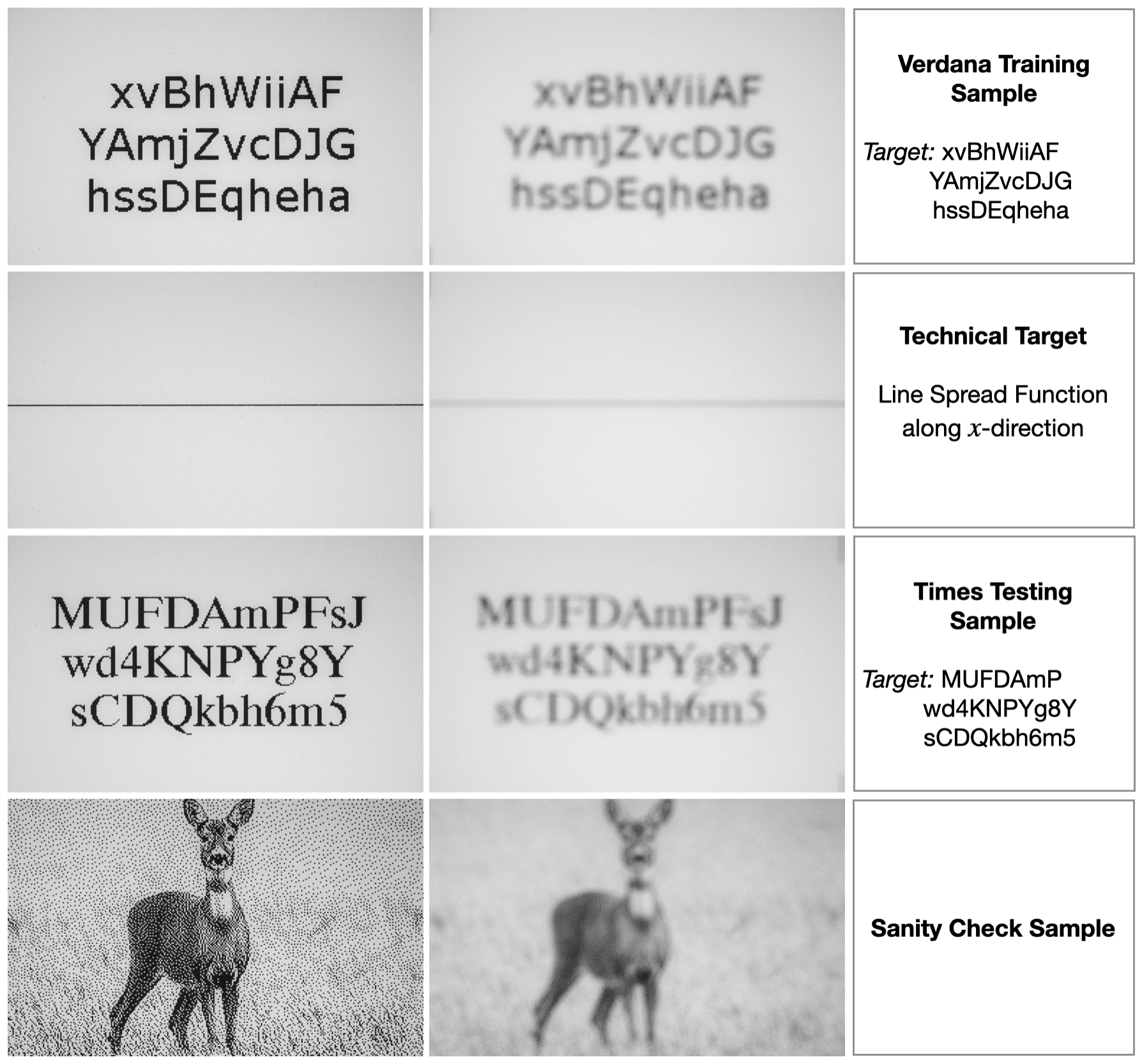}
	\caption{Examples of HDC training data pairs of sharp and blurred images showing text (top row) and a calibration target (second row). Examples of the test data (third row) and sanity check data (bottom row) were unknown to the participants before the submission deadline.}
	\label{fig:provided_data}
\end{figure}

The organizers of the HDC devised a carefully designed dataset that was divided into $20$ sub-problems of increasing defocus blur~\cite{HDC_Data}. Using two identical photo cameras observing an e-ink display through a beamsplitter mirror, they generated a high quality real-world dataset of paired sharp (ground truth) and blurred images. For this, one camera was always taking a sharp image (i.e., in-focus) while the other was put increasingly out-of-focus; see Fig.~\ref{fig:photo_setup} for the setup and Fig.~\ref{fig:provided_data} for examples of the resulting image pairs.

For each of the $20$ levels of increasing defocus blur, $203$ pairs of sharp and blurred images were generated and provided to the challenge participants. The $203$ images consist of $100$ samples with text showing the Verdana font and $100$ with text showing the Times New Roman font. On top of that, $3$ calibration images were provided showing a single central point, a single vertical line, and a single horizontal line, respectively; see Fig.~\ref{fig:provided_data} and Fig.~\ref{fig:background} (middle column). In addition to the image pairs, also the actual text string containing the displayed characters was provided for each of the $200$ text image examples.

\subsubsection{Objective}
A valid submission to the challenge is an algorithm that receives a blurred text image as input as well as the particular blur level to which the input belongs.\footnote{In other words, the participants were explicitly allowed to train $20$ separate deblurring neural networks for each of the $20$ levels, so that the algorithm can dispatch an input image to the network corresponding to its blur level.} The image is then to be deblurred by the submitted algorithm in order to make the displayed characters readable by an \emph{optical character recognition (OCR)} engine \cite{tesseract}.

\subsubsection{OCR Score}
The string predicted by the OCR engine after deblurring is then compared to the ground-truth text. The algorithm performance is quantified as a percentage score based on the Levenshtein distance~\cite{leven}. The Levenshtein distance captures the number of edits needed to transform one string into the other by means of insertions, deletions, or substitutions of characters. Following the challenge evaluation, only the central line of characters is considered. From now on, we refer to this performance measure as the \emph{OCR score}.

\subsubsection{Evaluation}
During the challenge evaluation, the algorithms have to \emph{clear} as many levels as possible, starting with the easiest (least amount of blur) and then proceeding to more and more difficult ones. An evaluation level is considered cleared, if the average OCR score exceeds $70\%$ on an unknown test dataset. 
This test data was generated similarly to the training data, but it was not disclosed to the participants before the submission deadline.
The test dataset differed from the training data in the set of text characters that were shown on the images: The training data only contains letters from the Latin alphabet while the test data also includes numerical digits, as shown in Fig.~\ref{fig:provided_data}.
The submitted algorithms were ranked by the number of \emph{consecutive} levels that were cleared and by a comparison of the average OCR scores on the highest cleared level in the case of a draw. 

\subsubsection{Sanity Check}
One additional restriction was made in the challenge rules regarding the ``generalization'' of the deblurring algorithms to \emph{non-text} images. Only general-purpose deblurring methods were allowed, disqualifying predictive models that would always result in text images regardless of the input image. This was ensured by a \emph{sanity check} testing the algorithms on different types of natural images, such as the one in Fig.~\ref{fig:provided_data}. However, this check was subjectively evaluated by the challenge organizers and the threshold for passing it was deliberately set rather low. It was \emph{\textquote{enough to have some sort of deconvolution effect, even a poor one, instead of always producing images of text}}~\cite{HDC_website}.

\subsubsection{Key Aspects}
The key difficulties arising with the HDC setup can be summarized as follows:
\begin{enumerate}
	\item The $20$ blur operators underlying the data for each level are unknown. In order to apply classical variational methods, an estimation of the forward operator is necessary.
	\item The $203$ samples provided for each level are insufficient for most modern data-driven solutions without heavy data augmentation or additional data generation.
	\item The real-world nature of the images can create unforeseen challenges, as no artificial simplification to the data can be made when modeling the forward operator.
\end{enumerate}
While the third point is of general relevance, the first problem is more specific to variational methods or hybrid approaches between model-based and data-driven techniques. On the other hand, the second aspect is an important limitation for (fully) data-driven solutions. However, both the first and second point can be tackled by carefully modeling the blur operators. 

\section{Methodology}\label{sec:meth}

In this section, we give a conceptual overview of our end-to-end deblurring pipeline as well as our data generation strategy. We establish the main aspects of our methodology for the HDC and motivate some of the key design choices. 
In the following, all steps are described for one of the $20$ challenge levels, and we proceed analogously for all other levels. 
More details on the concrete implementation, including the choices for different hyperparameters, can be found in our submitted GitHub repository~\cite{github_trippe}.

The centerpiece of our deblurring algorithm is an \emph{image-to-image} neural network $\mathcal{NN}[\theta]\colon\mathbb{R}^{mn}\to\mathbb{R}^{mn}$. Its weight parameters $\theta$ are determined via standard empirical risk minimization over the training dataset $(\mathbf{x}_i, \mathbf{y}_i)_{i=1}^N$:
\begin{equation*}
	{\arg \min}_\theta \, \frac{1}{N} \sum_{i=1}^N \| \mathcal{NN}[\theta]( \mathbf{y}_i) - \mathbf{x}_i \|_2^2 \ ,
\end{equation*}
where $\mathbf{y}_i \in \mathbb{R}^{mn}$ is a blurry input image and $\mathbf{x}_i \in \mathbb{R}^{mn}$ the sharp counterpart.
The specific architecture of $\mathcal{NN}$ is discussed below in Step~3. With only $N = 200 + 3$ data pairs provided per blur level, an immediate problem arising with this approach is the lack of an extensive dataset (see Section~\ref{sec:challenge}). Therefore, we aim at generating additional synthetic image pairs. However, this requires estimating the unknown forward operator in a first step.

\subsection{Step 1: Estimating the Forward Operator}\label{subsec:fwd}
As discussed in Section~\ref{sec:problem}, we model the forward blur process as a concatenation of a simple discrete convolution with a blur kernel $\mathbf{k}_B$ and a radial lens distortion $d[K_1,K_2]$, i.e., a discretization of \eqref{eq:cont_model_full} can be written as
\begin{equation*}
	\mathbf{y} = d[K_1,K_2](\mathbf{k}_B * \mathbf{x}).
\end{equation*}
Here, the distortion $d[K_1,K_2]$ denotes a discretized version of the previously introduced polynomial model \eqref{eq:distort} of order $Q=2$ with two radial distortion coefficients $K_1, K_2$, which seems sufficient when dealing with mild distortions~\cite{Wang}. The notation $d[K_1,K_2]$ highlights the dependency of the distortion on $K_1$ and $K_2$.

\begin{figure}
	\centering
	\begin{tabular}{ccc}
		& 
		&
		sharp bg removed 
		\\
		sharp image $\mathbf{x}$
		&
		sharp bg estimate $\mathbf{x}_0$
		&
		$\bar{\mathbf{x}} = \mathbf{x}-\mathbf{x}_0$
		\\
		\includegraphics[width=.25\linewidth]{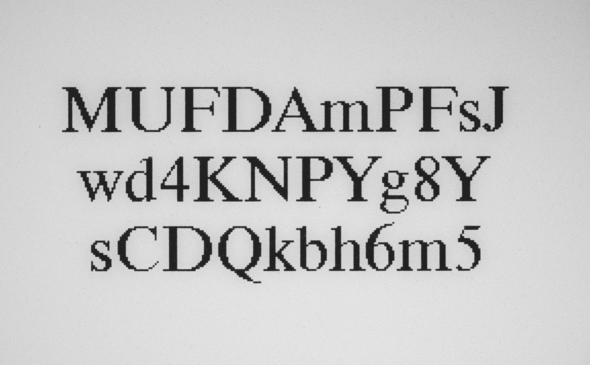}
		&
		\includegraphics[width=.25\linewidth]{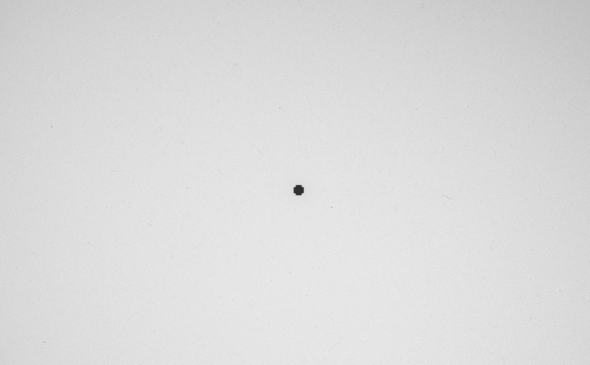}
		&
		\includegraphics[width=.25\linewidth]{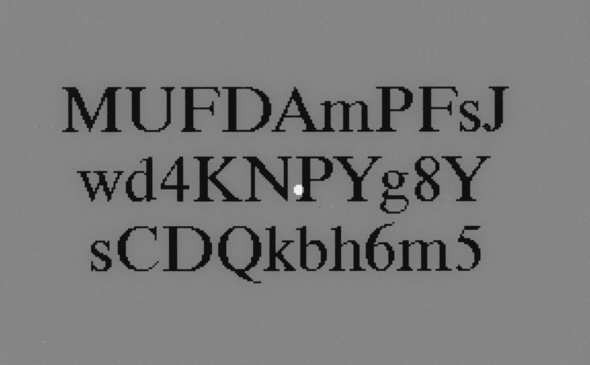}
		\\[.5em]
		blurred bg removed 
		&
		&
		blurred image 
		\\
		$\bar{\mathbf{y}} = \mathbf{k}_B * \bar{\mathbf{x}}$
		&
		blurred bg estimate $\mathbf{y}_0$
		& $\mathbf{y} = \bar{\mathbf{y}} + \mathbf{y}_0$
		\\
		\includegraphics[width=.25\linewidth]{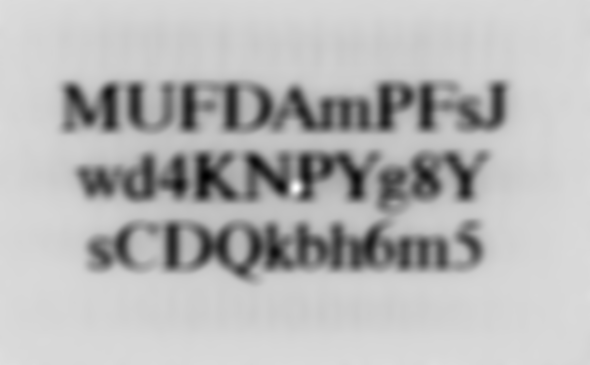}
		&
		\includegraphics[width=.25\linewidth]{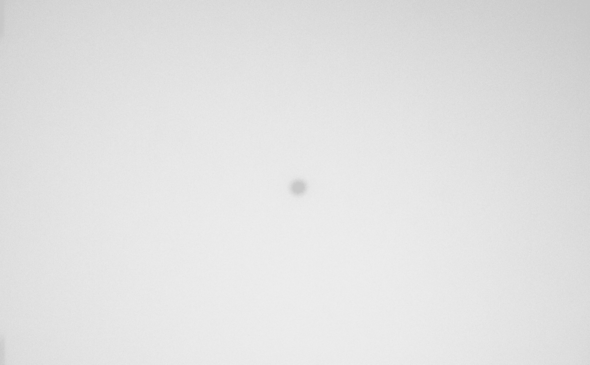}
		&
		\includegraphics[width=.25\linewidth]{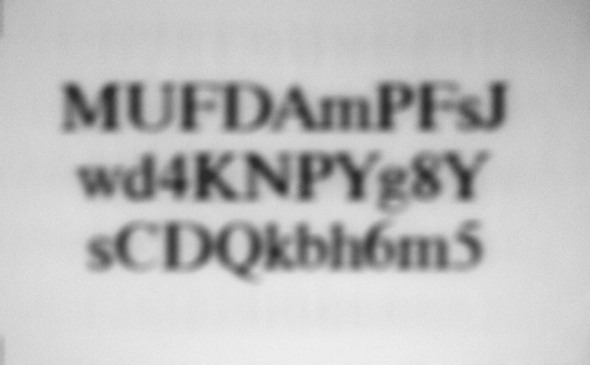}
	\end{tabular}
	\caption{Illustration of the forward blur model with background removal and addition. The background (bg) $\mathbf{x}_0$ and its corresponding blurred version $\mathbf{y}_0$ are estimated from the provided calibration target showing a single central point. They are subtracted and added before and after the discrete convolution with the estimated blur kernel, respectively.}
	\label{fig:background}
\end{figure}

The parameters $\mathbf{k}_B, K_1, K_2$ are estimated based on the provided training data $(\mathbf{x}_i, \mathbf{y}_i)_{i=1}^N$ by solving the minimization problem
\begin{equation}\label{eq:op_min}
	{\arg \min}_{\mathbf{k}_B, K_1, K_2} \, \frac{1}{N} \sum_{i=1}^{N} \| d[K_1, K_2] (\mathbf{k}_B * \mathbf{\bar{x}}_i) - \mathbf{\bar{y}}_i \|_2^2 \ .
\end{equation}
Note that for a robust estimation of the parameters in this step, it was essential to use slightly modified data pairs $(\bar{\mathbf{x}}_i,\bar{\mathbf{y}}_i) = (\mathbf{x}_i-\mathbf{x}_0,\mathbf{y}_i-\mathbf{y}_0)$ in which we approximately remove the image background. As sharp and blurred background approximation $\mathbf{x}_0$ and $\mathbf{y}_0$, we take the calibration target image showing a single central point and its blurred version (corresponding to a point spread function (PSF)), respectively; see also Fig.~\ref{fig:background}.

\begin{figure}
	\centering
	\begin{tabular}{cccc}
		level 4 & level 9 & level 14 & level 19 \\
		\includegraphics[width=2.25cm]{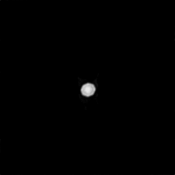} & 
		\includegraphics[width=2.25cm]{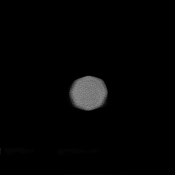} & 
		\includegraphics[width=2.25cm]{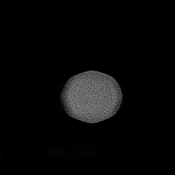} & 
		\includegraphics[width=2.25cm]{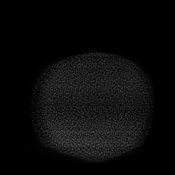}      
	\end{tabular}
	\caption{Estimated kernels $\mathbf{k}_B$ for blur levels $4$, $9$, $14$, and $19$ with enhanced contrast. The octagonal shape is typical for the polygonal shutter lenses of modern apertures, indicating that these estimated kernels reflect the underlying physical reality.}
	\label{fig:our_kernels}
\end{figure}

We found that a kernel size of $s=701$ for $\mathbf{k}_B\in\mathbb{R}^{s\times s}$ is large enough to accurately capture the spatial extent of all 20 blur processes.
We solve \eqref{eq:op_min} through stochastic gradient descent with the Adam optimizer~\cite{adam} and PyTorch's automatic differentiation \cite{pas+17} for computing the gradients.
A similar approach was recently taken in the context of computed tomography to estimate the unknown fanbeam geometry of a Radon transform~\cite{AAPM}.
Some examples of the estimated kernels $\mathbf{k}_B$ can be found in Fig.~\ref{fig:our_kernels}.

\subsection{Step 2: Generating a Data Stream}\label{step 2}
\begin{figure}
	\centering
	\begin{tabular}{cc@{\,}c}
		(a) & \rotatebox[origin=c]{90}{original HDC} & \includegraphics[valign=c,width=0.6\linewidth]{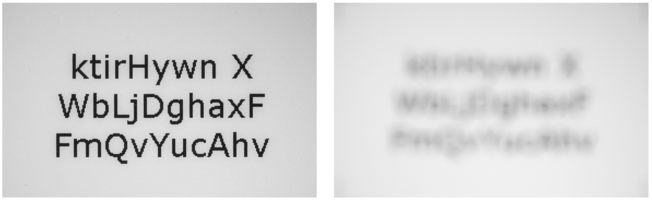} \\
		(b) & \rotatebox[origin=c]{90}{synthetic HDC} & \includegraphics[valign=c,width=0.6\linewidth]{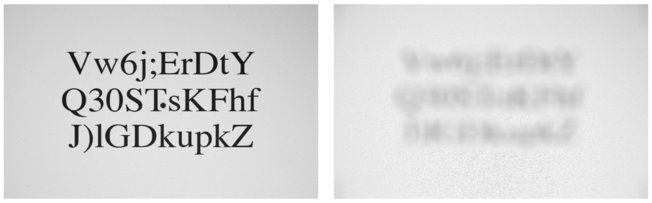} \\
		(c) & \rotatebox[origin=c]{90}{sanity check} & \includegraphics[valign=c,width=0.6\linewidth]{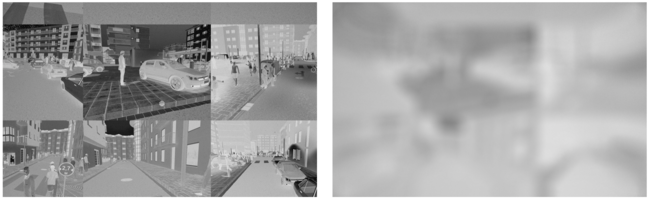}
	\end{tabular}
	\caption{Examples of sharp and blurred image pairs from the (a) original HDC data, (b) synthetic HDC data, and (c) synthetic sanity check data. All blurred images correspond to the same blur level.}
	\label{fig:training_data}
\end{figure}

With the estimated parameters $\mathbf{k}_B$, $K_1$, $K_2$ from Step~1, the generation of additional data becomes possible. We synthesize two different types of samples, which leads to three sample categories during training:

\begin{enumerate}
	\item \emph{HDC Data:} The original dataset provided by the organizers (see Fig.~\ref{fig:training_data}(a)).
	\item \emph{Synthetic HDC Data:} Aiming to imitate the HDC dataset, these samples are generated using random text strings to create synthetic sharp images $\mathbf{x}$. These are blurred and distorted using the estimated forward model from Step 1 to obtain $\widetilde{\mathbf{y}}$. Finally, in order to account for measurement noise, we add amplitude-dependent Gaussian noise to obtain $\mathbf{y} = \widetilde{\mathbf{y}} + \widetilde{\mathbf{y}} \odot \boldsymbol{\eta}$, where $\boldsymbol{\eta} \sim \mathcal{N}(\mathbf{0}, 0.001\cdot\mathbf{I})$.
	The synthetic data distribution differs from the original HDC data by design, as we include more diverse fonts and include numbers as well as some special characters in the randomly generated strings (see Fig.~\ref{fig:training_data}(b)).
	\item \emph{Synthetic Sanity Data:} We generate another dataset which intentionally and more drastically differs from the HDC data.
	For that purpose, we use the Synthia dataset~\cite{synthia}, providing $13407$ samples of synthetic traffic images, to generate out-of-distribution data. These images more closely resemble natural images and are used to ensure that we pass the sanity check.
	We perform data augmentation on these samples by patching $9$ randomly selected Synthia samples in a $3 \times 3$ grid and then randomly cropping the resulting image to the same width and height as the original HDC data. As before, the corresponding blurred images are obtained using the estimated forward model from Step~1 (see Fig.~\ref{fig:training_data}(c)).
\end{enumerate}

\subsection{Step 3: Network Architecture}\label{sec:step_3}   
Similarly to our estimated forward model, the reconstruction model can be divided into two steps. We start with a pre-processing step that aims at approximately inverting the radial distortion. For this we use the division model \eqref{eq:undistort} for \emph{distortion correction} and denote its discretization by $u[K_1,K_2]$, with the radial distortion parameters $K_1,K_2$ as in Step~1. This pre-processing step is followed by a slightly modified \emph{U-Net} $\mathcal{U}$~\cite{U-Net} resulting in the overall network model
\begin{equation*}
	\mathcal{NN}[\theta,K_1,K_2]\colon\mathbb{R}^{mn}\to\mathbb{R}^{mn}, \ \mathbf{x}\mapsto \left(\mathcal{U}[\theta]\circ u[K_1,K_2]\right)(\mathbf{x}).
\end{equation*}
The standard U-Net is a fully convolutional neural network architecture that has been proven very effective in solving many types of (image-to-image) reconstruction problems, e.g., see~\cite{AAPM, robustness, hsqdsr19}. Since fully convolutional architectures are primarily designed to model local dependencies between pixels, the pre-processing step seems natural to capture global effects due to the spatial variance in the deblurring problem.
The main modification compared to the standard U-Net amounts to increasing the number of down- and up-sampling convolutional levels (thus increasing the depth of the ``U'') in order to increase the receptive field\footnote{In a two dimensional CNN, the receptive field of a neuron in any given layer are those pixels in the input image that can contribute to the neurons activation. For standard CNNs, the receptive field is a rectangular patch of the input.} in the lowest bottleneck layers \cite{rec_field}. Following the interpretation of the U-Net as an auto-encoder, we therefore ensure that a single neuron in the bottleneck layer can ``see'' an input image patch large enough to capture the $701 \times 701$ blur kernel plus some extra space around it. Instead of the common $4$ down-sampling levels, we therefore use $6$ levels and an additional introductory pooling layer.
Fig.~\ref{fig:pipeline} depicts the two reconstruction steps (purple arrow) and schematically describes the modified U-Net architecture (bottom).

\subsection{Step 4: Pre-training, Fine-tuning, and Validation}\label{sec:step_4}
Using the data generation pipeline from Step~2, we train the reconstruction model from Step~3 in two phases, followed by a validation phase:
\begin{enumerate}
	\item \emph{Pre-training:} In a first stage, the network is pre-trained using only synthetic data. The samples are generated in a data stream and spread out over $150$ epochs of training.
	In total, $150000$ synthetic HDC samples are used and a smaller portion of $7500$ synthetic sanity check images.
	This ratio is chosen to nudge the network towards being a general-purpose deblurring algorithm while focusing on the actual goal of deblurring text images.
	\item \emph{Fine-tuning:} In the second stage, we reset the optimizer parameters and fine-tune the network in $350$ training epochs, now focusing on the original HDC data: we use $178$ of the provided $200$ samples per blur level and an additional $40$ (randomly generated) synthetic HDC samples as well as $5$ (randomly generated) sanity check images.
	\item \emph{Validation:} The validation after each epoch is performed on $20$ of the remaining original HDC samples.
	The best epoch per blur level is chosen according to the highest OCR score on these samples. To test the performance of our final reconstruction models, the remaining $2$ original HDC samples were used.
\end{enumerate}
Note that we still use a small amount of synthetic data in the fine-tuning stage to avoid catastrophic forgetting and retain the generalization capacities of the network acquired in the pre-training stage. 
Besides passing the sanity check, this is particularly useful for the final evaluation, which is performed on a similar but not identical data distribution as the  HDC training data.

\section{Results}\label{sec:res}
This section presents the main results of our work, focusing on qualitative visual inspections and quantitative assessments through the OCR score.
We begin with a discussion of the challenge results (Section~\ref{subsec:results:chall}). We then proceed with findings related to the forward operator estimation (Section~\ref{subsec:results:fwd}) and finally present ablation studies underpinning several design choices and aspects of our overall reconstruction pipeline (Section~\ref{subsec:results:reconstr}).
All examples are produced with test data, which consists of $N = 40$ samples per blur level and was not seen by any of the networks in the training phase.
We only show exemplary results for the blur levels  $4$, $9$, $14$, and $19$ (an exception is the last ablation study where we show level $17$). 
Apart from that, we mostly show text in Times New Roman font, since it was more challenging due to finer details in the serifs.

\subsection{Challenge Results} \label{subsec:results:chall}
\begin{figure}
	\centering
	\begin{tabular}{c@{\,}c@{\,}c@{\,}c@{\,}c}
		\textbf{\ level \ } & sharp image & blurry image & reconstruction & metrics\\
		\vspace{0.15em}
		
		\textbf{4} & 
		\includegraphics[valign=c,width=2.7cm]{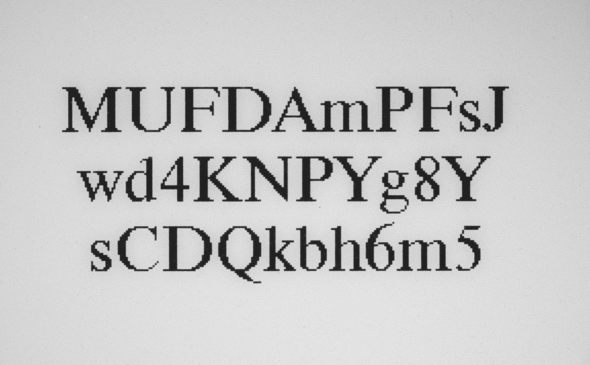} & 
		\includegraphics[valign=c,width=2.7cm]{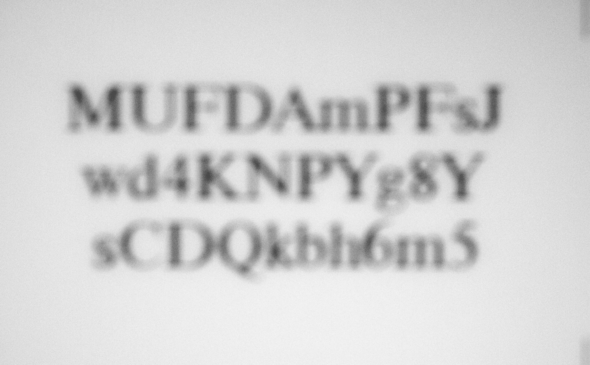} & 
		\includegraphics[valign=c,width=2.7cm]{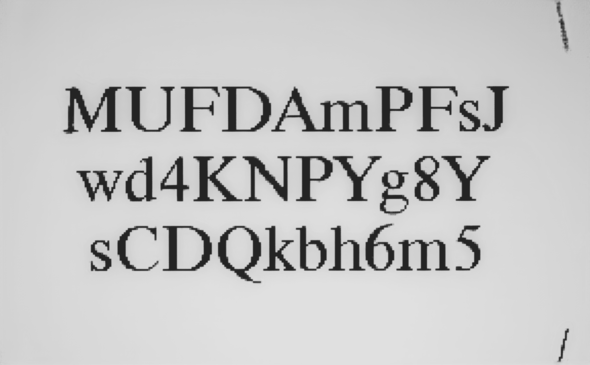} &
		\includegraphics[valign=c,width=1.85cm]{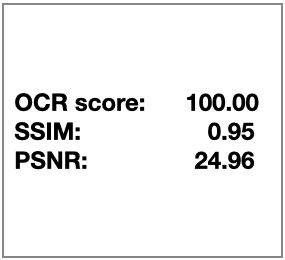}
		\\
		\vspace{0.15em}
		
		\textbf{9} & 
		\includegraphics[valign=c,width=2.7cm]{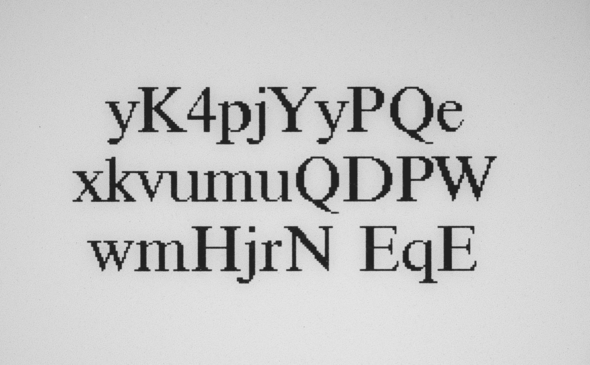} & 
		\includegraphics[valign=c,width=2.7cm]{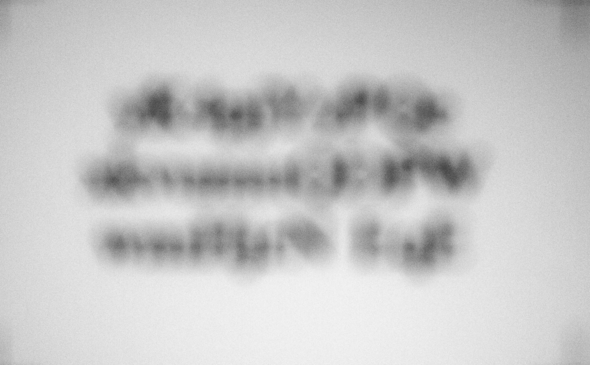} & 
		\includegraphics[valign=c,width=2.7cm]{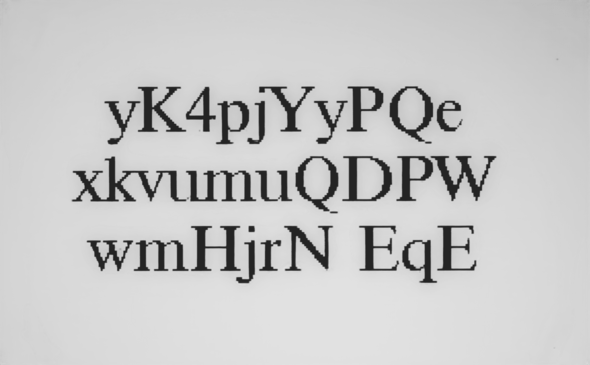} &
		\includegraphics[valign=c,width=1.85cm]{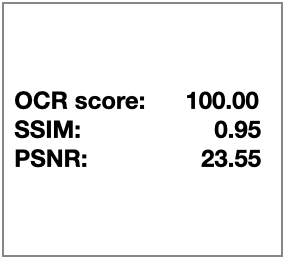}
		\\
		\vspace{0.15em}
		
		\textbf{14} & 
		\includegraphics[valign=c,width=2.7cm]{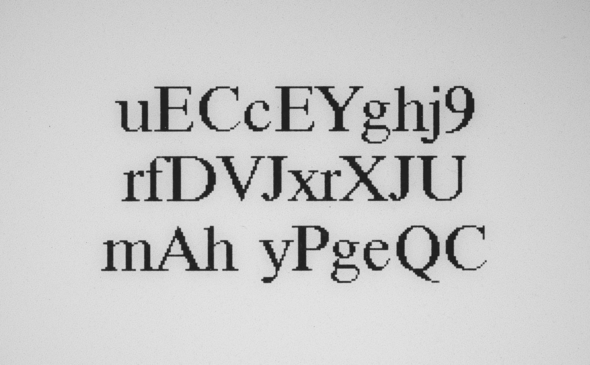} & 
		\includegraphics[valign=c,width=2.7cm]{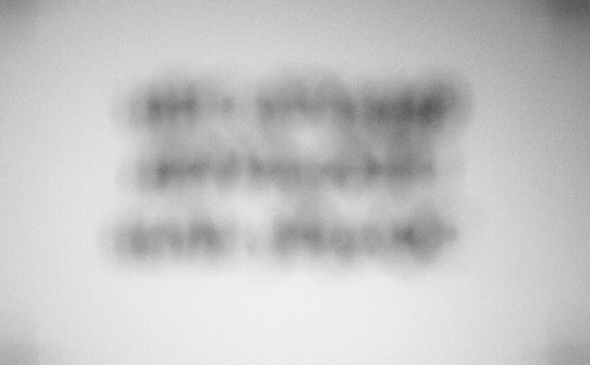} & 
		\includegraphics[valign=c,width=2.7cm]{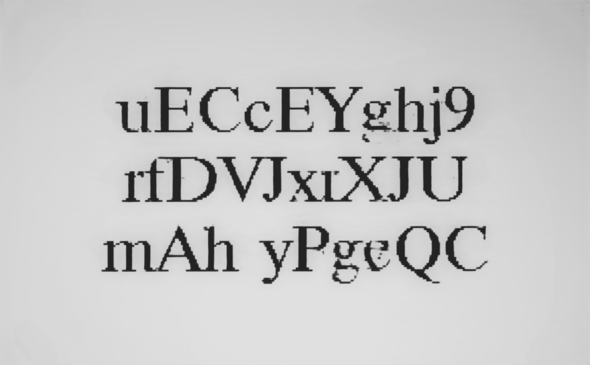} &
		\includegraphics[valign=c,width=1.85cm]{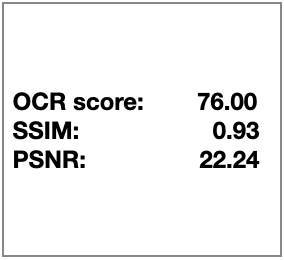}
		\\
		\vspace{0.15em}
		
		\textbf{14} & 
		\includegraphics[valign=c,width=2.7cm]{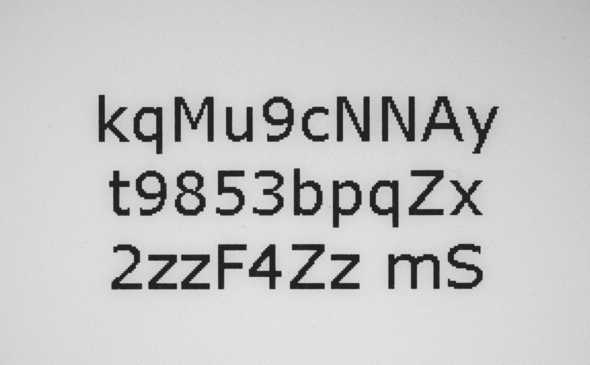} & 
		\includegraphics[valign=c,width=2.7cm]{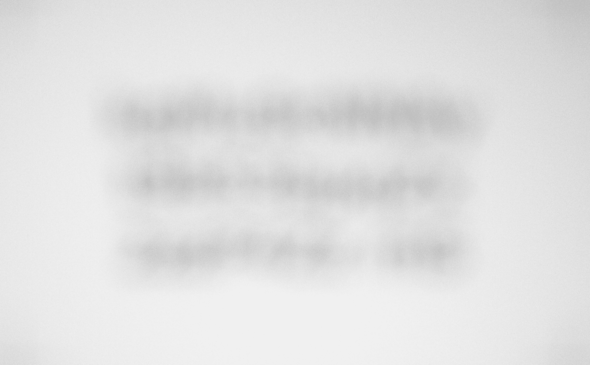} & 
		\includegraphics[valign=c,width=2.7cm]{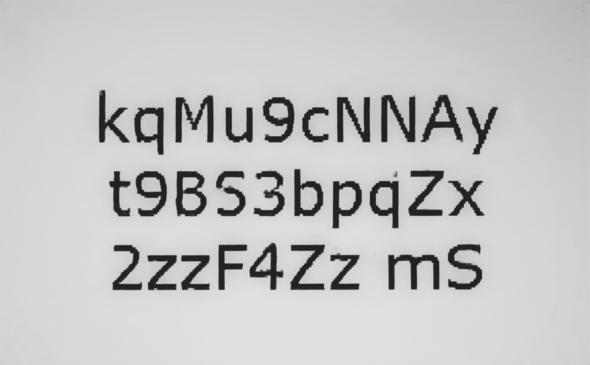} &
		\includegraphics[valign=c,width=1.85cm]{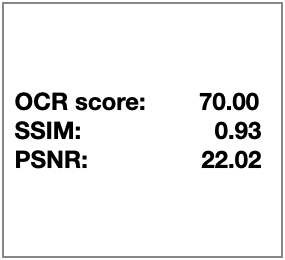}
		\\      
		\vspace{0.15em}
		
		\textbf{19} & 
		\includegraphics[valign=c,width=2.7cm]{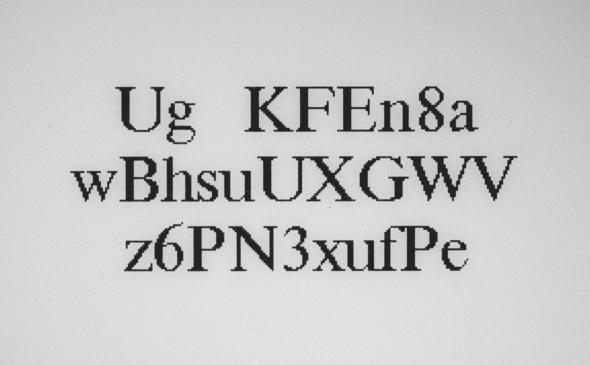}& 
		\includegraphics[valign=c,width=2.7cm]{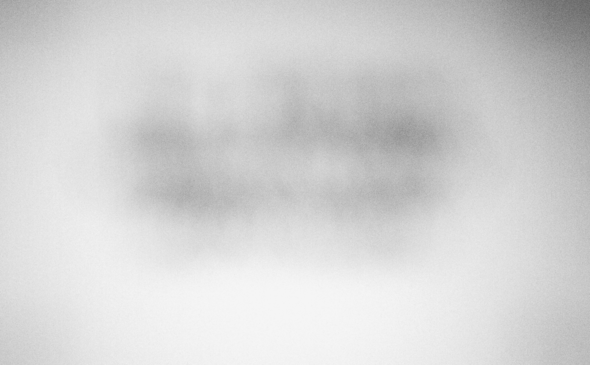} & 
		\includegraphics[valign=c,width=2.7cm]{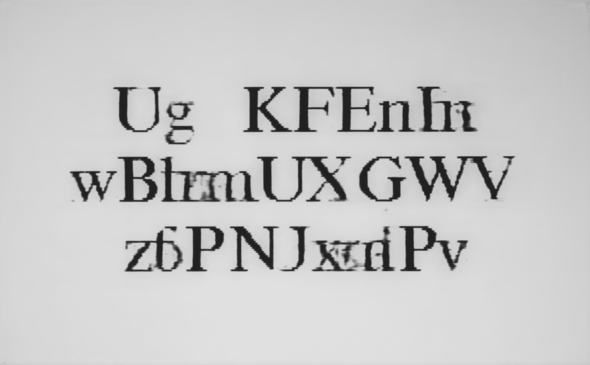} &
		\includegraphics[valign=c,width=1.85cm]{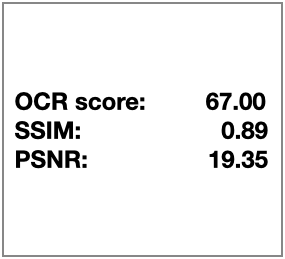}
		\\
		\vspace{0.15em}
		
		\textbf{19} & 
		\includegraphics[valign=c,width=2.7cm]{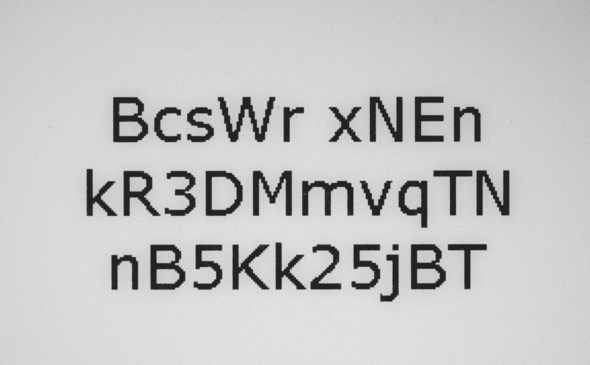} & 
		\includegraphics[valign=c,width=2.7cm]{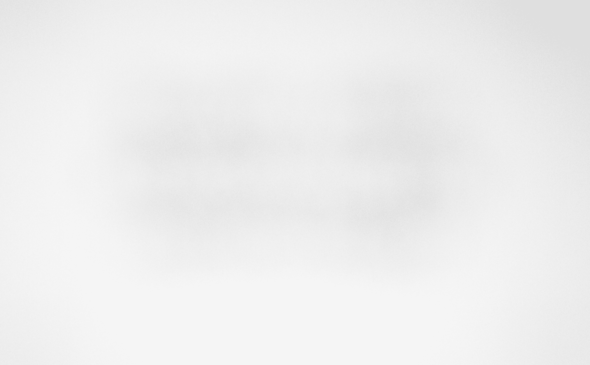} & 
		\includegraphics[valign=c,width=2.7cm]{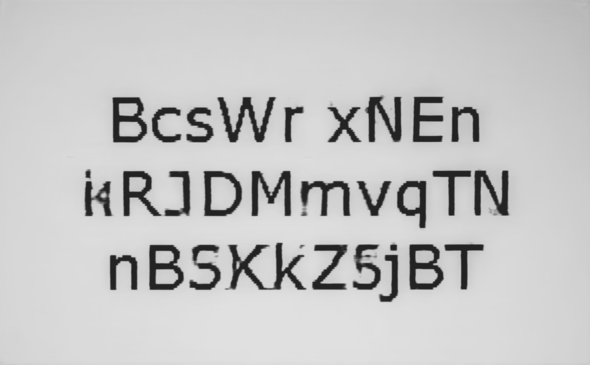} &
		\includegraphics[valign=c,width=1.85cm]{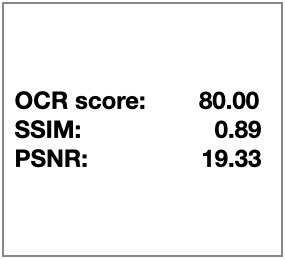}
	\end{tabular}
	\caption{Example reconstruction results (right column) together with the sharp ground-truth images (left column) and blurry input images (middle column) for the blur levels $4$, $9$, $14$, and $19$. For the levels $14$ and $19$, both fonts (upper: Times New Roman, lower: Verdana) are shown. The right column shows the OCR Score for these particular samples, and for the sake of completeness, also the standard evaluation metrics SSIM and PSNR are reported; see Fig.~\ref{fig:ablation_avg_ocr}, Fig.~\ref{fig:ssim_avg} and Fig.~\ref{fig:psnr_avg} for corresponding average scores.}
	\label{fig:final_results_test}
\end{figure}

\begin{figure}
	\centering
	\begin{tabular}{c@{\,}c@{\,}c@{\,}c@{\,}c}
		\textbf{\ level \ } & blurry image & reconstruction & blurry image & reconstruction \\
		\vspace{0.15em}
		
		\textbf{4} & 
		\includegraphics[valign=c,width=2.7cm]{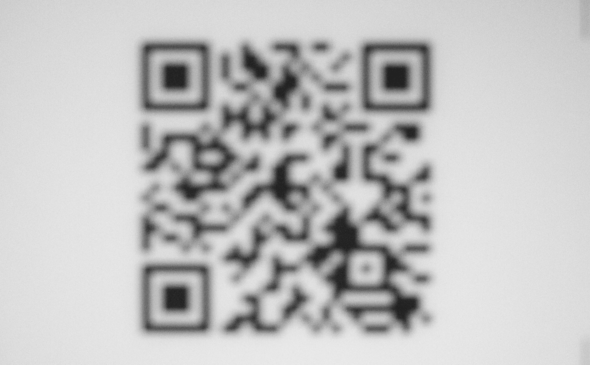} & 
		\includegraphics[valign=c,width=2.7cm]{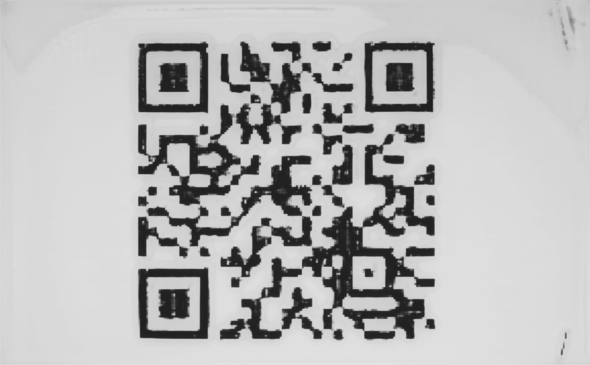} & 
		\includegraphics[valign=c,width=2.7cm]{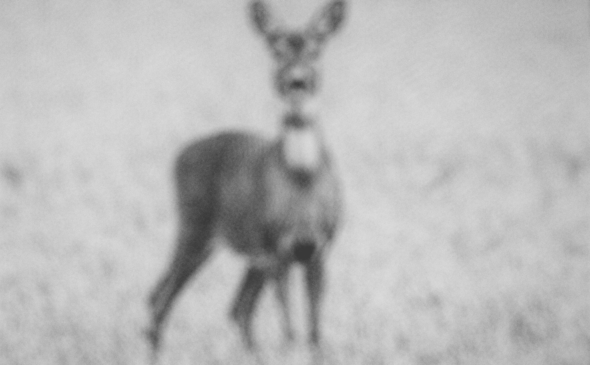} & 
		\includegraphics[valign=c,width=2.7cm]{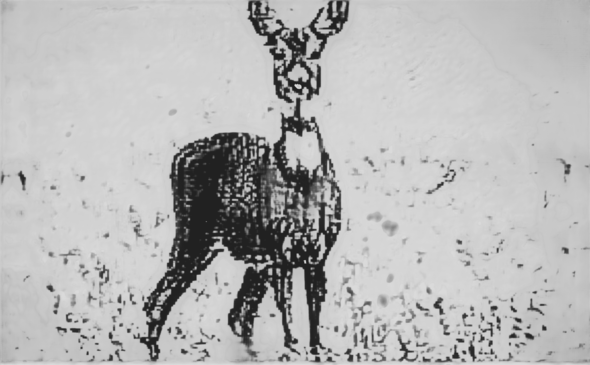} \\     
		\vspace{0.15em}
		
		\textbf{9} & 
		\includegraphics[valign=c,width=2.7cm]{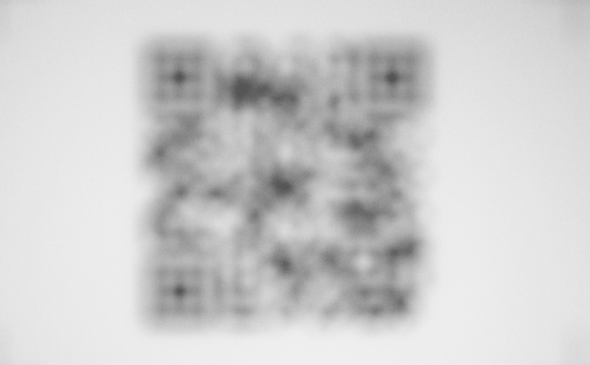} & 
		\includegraphics[valign=c,width=2.7cm]{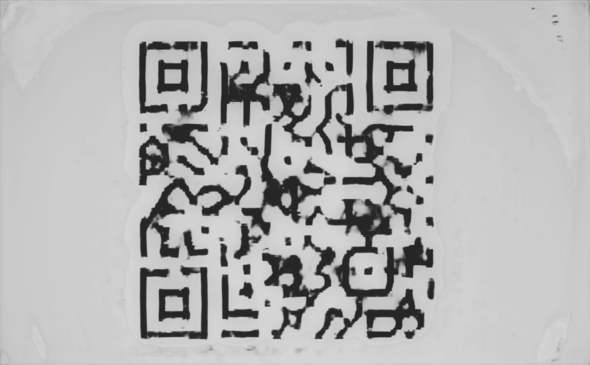} & 
		\includegraphics[valign=c,width=2.7cm]{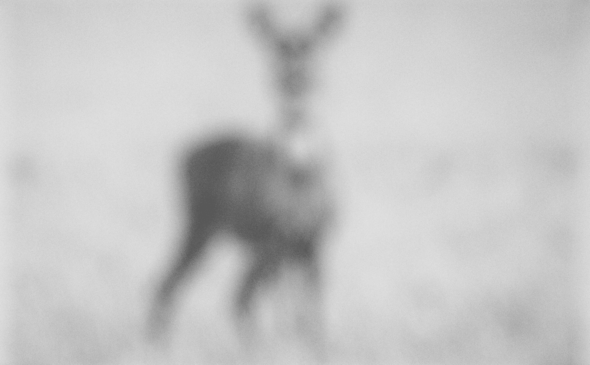} & 
		\includegraphics[valign=c,width=2.7cm]{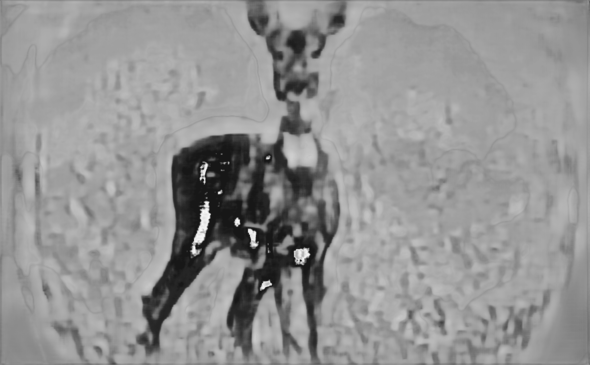} \\      
		\vspace{0.15em}
		
		\textbf{14} & 
		\includegraphics[valign=c,width=2.7cm]{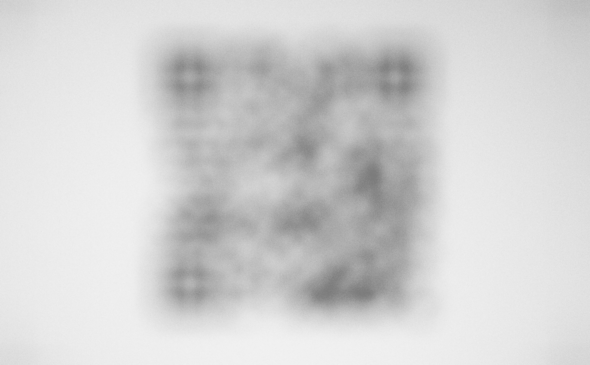} & 
		\includegraphics[valign=c,width=2.7cm]{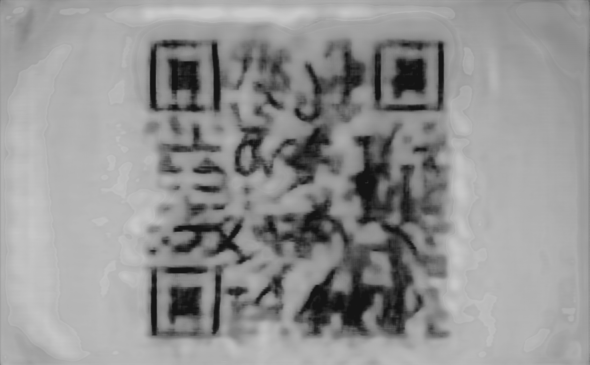} & 
		\includegraphics[valign=c,width=2.7cm]{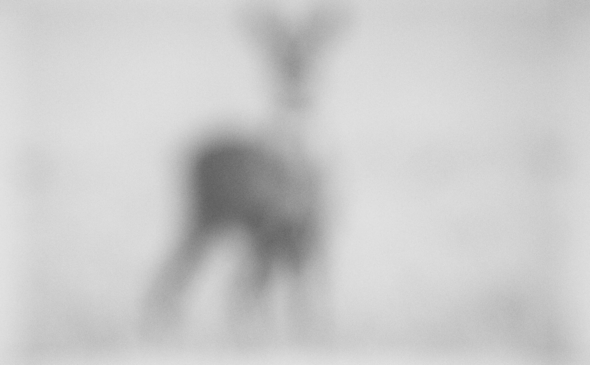} & 
		\includegraphics[valign=c,width=2.7cm]{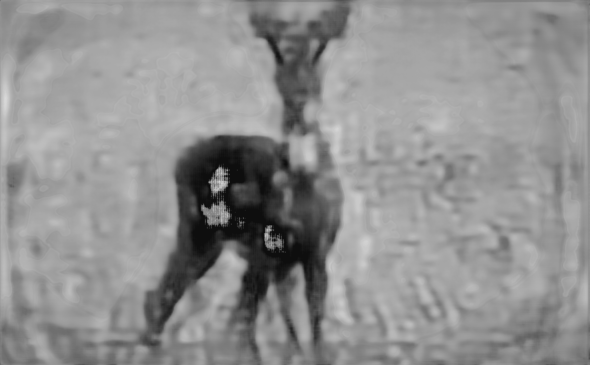} \\      
		\vspace{0.15em}
		
		\textbf{19} & 
		\includegraphics[valign=c,width=2.7cm]{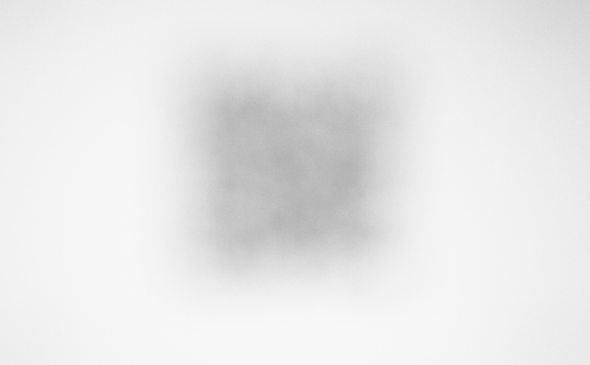} & 
		\includegraphics[valign=c,width=2.7cm]{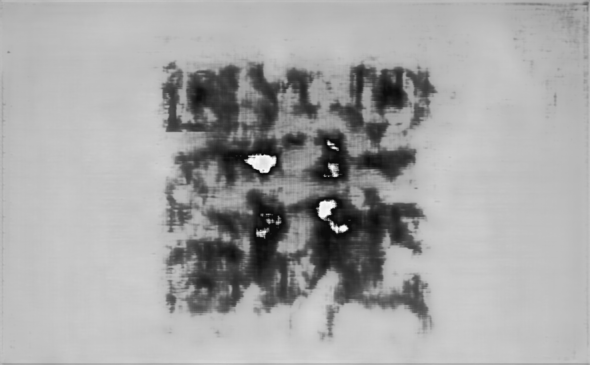} & 
		\includegraphics[valign=c,width=2.7cm]{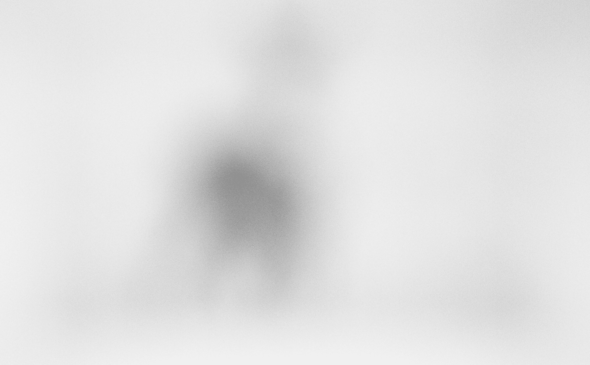} & 
		\includegraphics[valign=c,width=2.7cm]{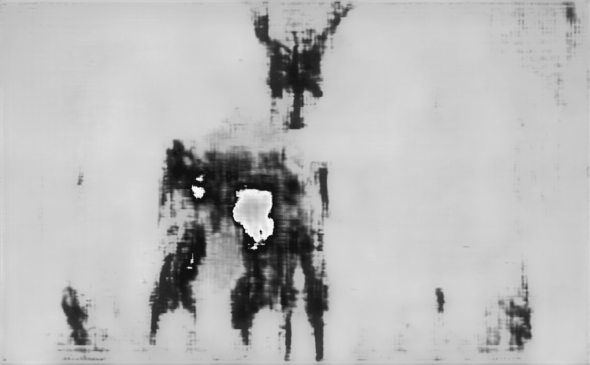}      
	\end{tabular}
	\caption{Example reconstructions of two out-of-distribution images from the HDC sanity check data, shown for blur levels $4$, $9$, $14$, and $19$.}
	\label{fig:sanity}
\end{figure}

\begin{figure}
	\centering
	\includegraphics[width=.8\linewidth]{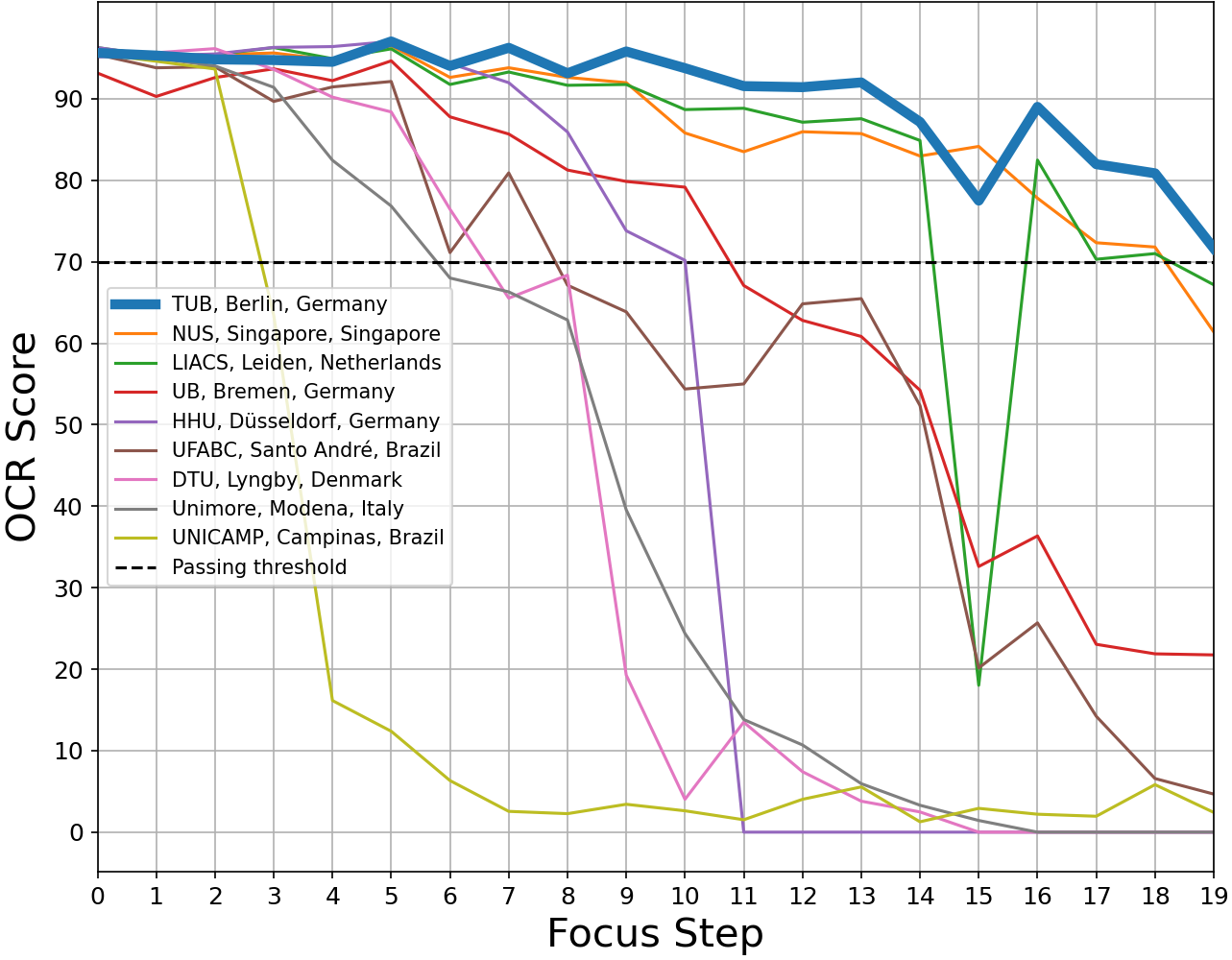}  
	\caption{Summary of results from the HDC. The figure plots the blur level against the average OCR scores achieved by each participating team on the HDC test set. Our winning submission is highlighted as the bold line. Note that some teams have submitted multiple methods, and we took the most accurate one in each case.}
	\label{fig:hdc_res}
\end{figure}

Examples of reconstruction results from the challenge test set are shown in Fig.~\ref{fig:final_results_test}.
Similarly, some challenge sanity check images can be found in Fig.~\ref{fig:sanity}. 
For more examples, we refer to the official HDC website~\cite{HDC_results}.
The average OCR score of our deblurring algorithm on the challenge test set is shown in Fig.~\ref{fig:hdc_res} and compared to the corresponding OCR scores of all competing teams. Our submission is the only one that constantly achieves an OCR score above the passing threshold of $70\%$, thereby clearing all challenge levels.

\subsection{Forward Operator}\label{subsec:results:fwd}
In order to better understand the impact of the radial lens distortion in the forward model and the quality of the approximate inverse distortion, we have conducted the following two experiments.

\begin{figure}
	\centering
	\begin{tabular}{@{}c@{\,}c@{\,\,\,}c@{\,}c@{\,\,\,}c@{\,}c@{}}
		\textbf{\ level \ } & original HDC & only conv. & residual & conv. \& dist.\ & residual \\
		\vspace{0.15em}		
		\textbf{4} &
		\includegraphics[valign=c,width=2.15cm]{img/fwd/naive/gt/gt_step_04.png} &
		\includegraphics[valign=c,width=2.15cm]{img/fwd/naive/final_step_04.png} &
		\includegraphics[valign=c,width=2.15cm]{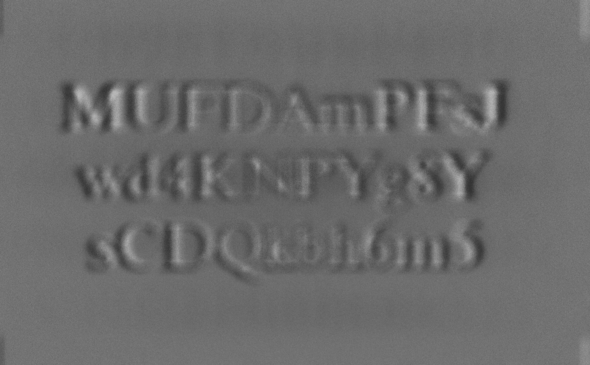} &
		\includegraphics[valign=c,width=2.15cm]{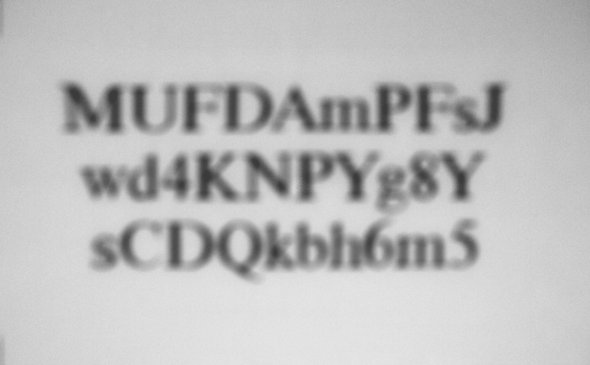} &
		\includegraphics[valign=c,width=2.15cm]{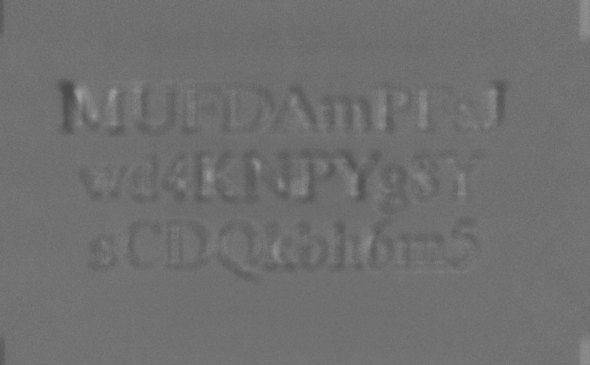}
		\\
		\vspace{0.15em}
		
		\textbf{9} &
		\includegraphics[valign=c,width=2.15cm]{img/fwd/naive/gt/gt_step_09.png} &
		\includegraphics[valign=c,width=2.15cm]{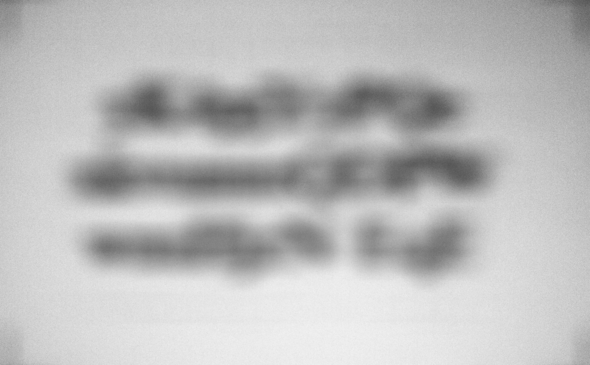} &
		\includegraphics[valign=c,width=2.15cm]{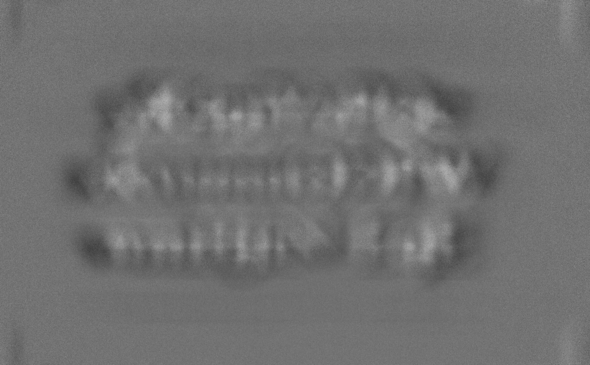} &
		\includegraphics[valign=c,width=2.15cm]{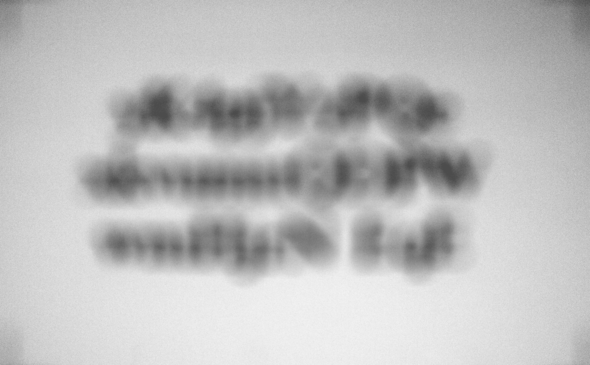} &
		\includegraphics[valign=c,width=2.15cm]{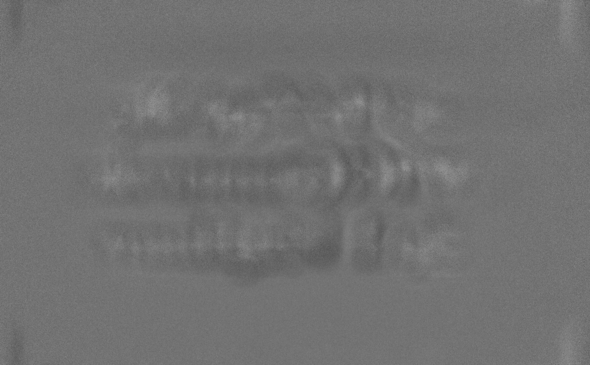} \\
		\vspace{0.15em}
		
		\textbf{14} &
		\includegraphics[valign=c,width=2.15cm]{img/fwd/naive/gt/gt_step_14.png} &
		\includegraphics[valign=c,width=2.15cm]{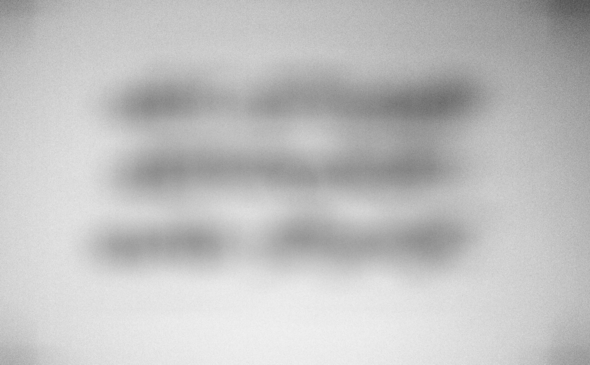} &
		\includegraphics[valign=c,width=2.15cm]{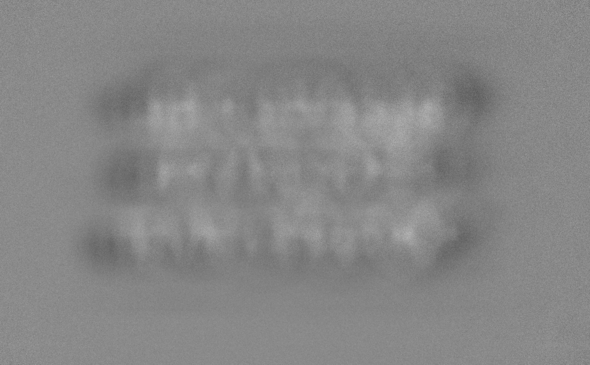} &
		\includegraphics[valign=c,width=2.15cm]{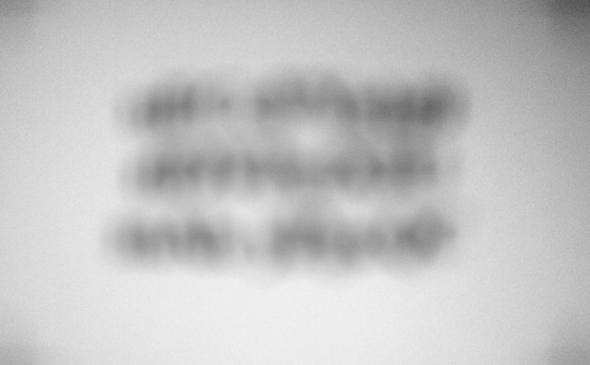} &
		\includegraphics[valign=c,width=2.15cm]{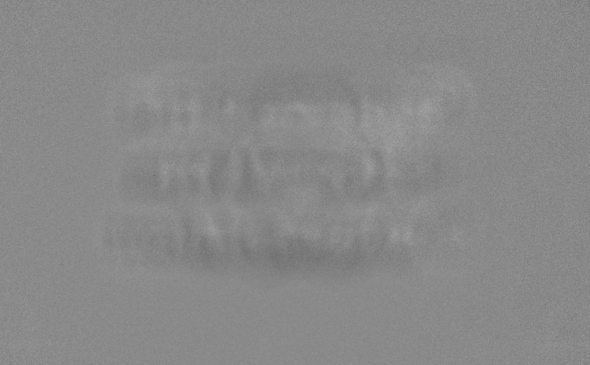} \\
		\vspace{0.15em}
		
		\textbf{19} &
		\includegraphics[valign=c,width=2.15cm]{img/fwd/naive/gt/gt_step_19.png} &
		\includegraphics[valign=c,width=2.15cm]{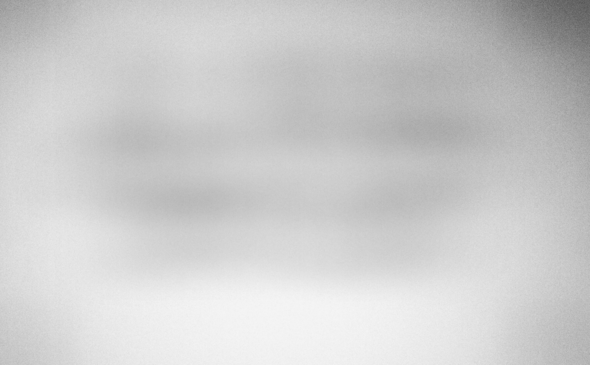} &
		\includegraphics[valign=c,width=2.15cm]{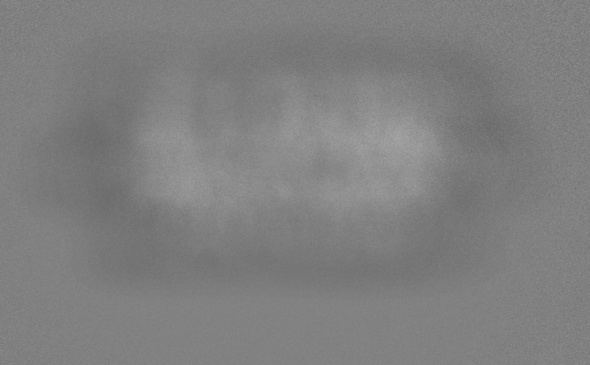} &
		\includegraphics[valign=c,width=2.15cm]{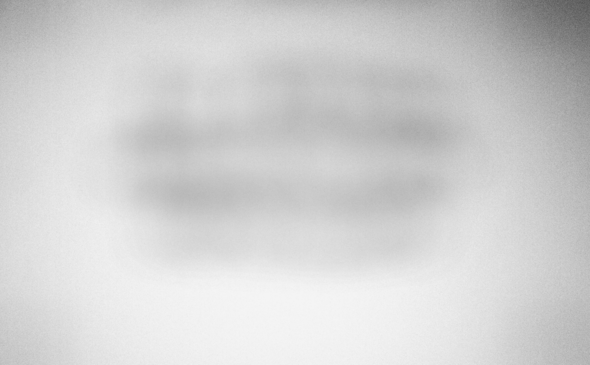} &
		\includegraphics[valign=c,width=2.15cm]{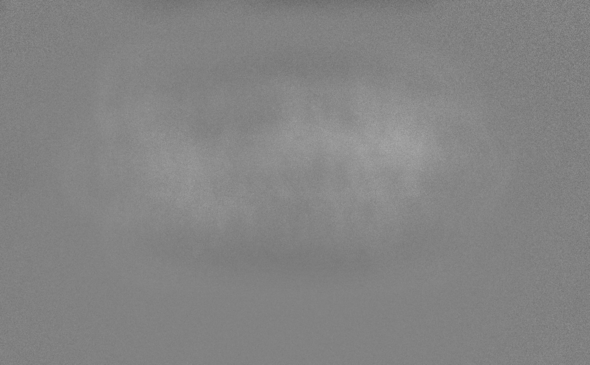} \\
	\end{tabular}
	\caption{Comparison of a blurry image from the HDC dataset with the results of two simulated forward models: a simple one, using only a convolution with a single spatially invariant blur kernel, and the one from Section~\ref{subsec:fwd}, using a spatially variant blur kernel that accounts for radial lens distortion.}
	\label{fig:fwd_compare}
\end{figure}

\begin{figure}
	\centering
	\begin{tabular}{c@{\,\,}c@{\,\,\,\,}c}
		\textbf{\ level \ } & distorted & inverted \\
		
		\textbf{4} & 
		\includegraphics[valign=c,width=3.3cm]{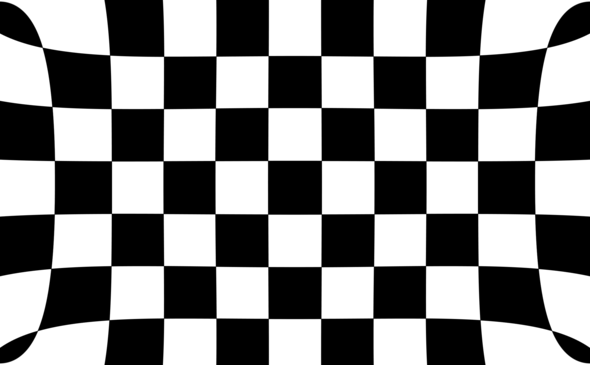} &
		\includegraphics[valign=c,width=3.3cm]{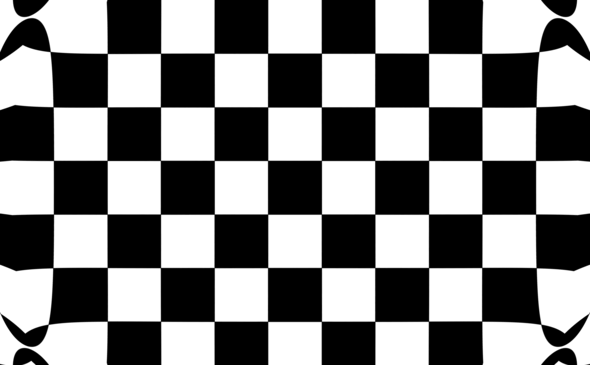} \\[3em]
		
		\textbf{9} & 
		\includegraphics[valign=c,width=3.3cm]{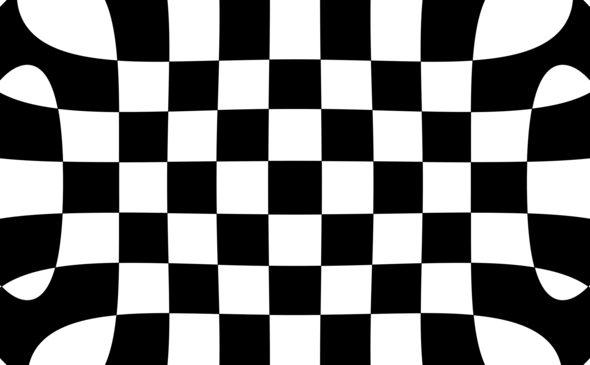} &
		\includegraphics[valign=c,width=3.3cm]{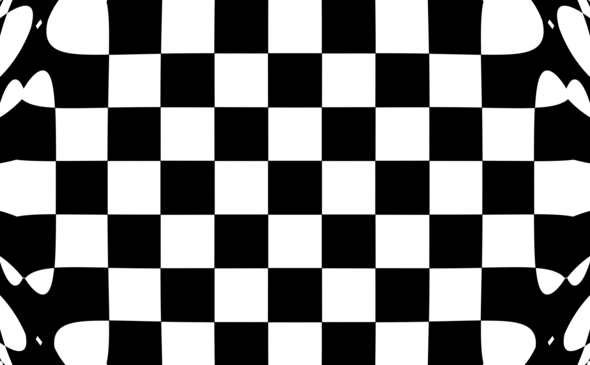} \\[3em]
		
		\textbf{14} & 
		\includegraphics[valign=c,width=3.3cm]{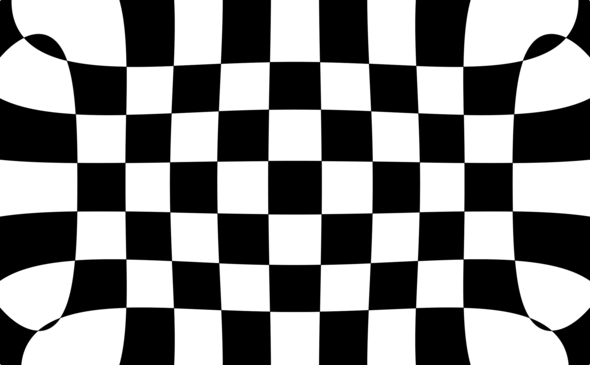} &
		\includegraphics[valign=c,width=3.3cm]{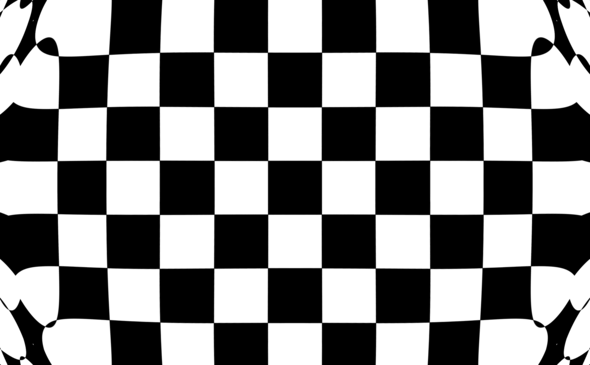} \\[3em]
		
		\textbf{19} & 
		\includegraphics[valign=c,width=3.3cm]{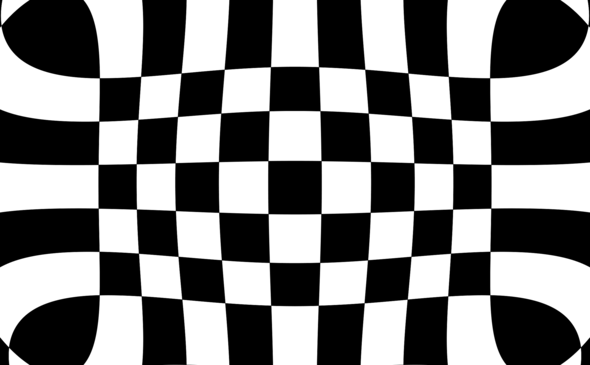} &
		\includegraphics[valign=c,width=3.3cm]{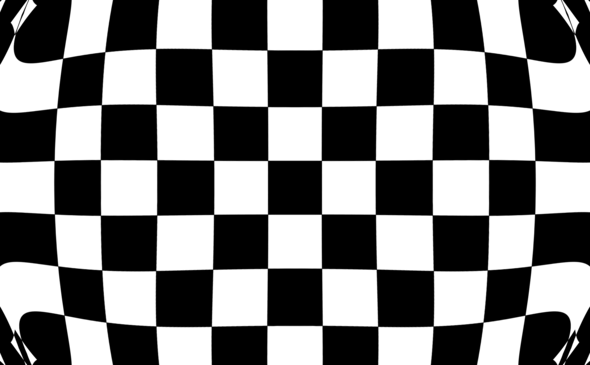} 
	\end{tabular}
	\caption{Visualization of the estimated radial lens distortion and the corresponding inverse distortion, shown for the blur levels $4$, $9$, $14$, and $19$.}
	\label{fig:dist_chess}
\end{figure}

\subsubsection{Ablation Study: Radial Distortion}
Here, we compare our estimated forward model (including a radial lens distortion) with the simpler one, which only estimates a spatially invariant blur kernel (convolution without any distortion). Fig.~\ref{fig:fwd_compare} displays an original blurry image from the HDC dataset as well as the artificially blurred images generated by the estimated forward models. For the sake of better visibility, we also show the respective residuals. The contrast has been enhanced for all residual images, but the amount of enhancement is constant within each blur level for better comparability.
Most notably, the missing radial distortion can be identified in the residuals of the simplified blur model in the form of a radially symmetric shadow pointing outwards.

\subsubsection{Inversion of the Radial Lens Distortion}\label{sec:res_distort_undistort}
Using a simple $7 \times 11$ chessboard pattern for visualization purposes, Fig.~\ref{fig:dist_chess} showcases the severity of the estimated radial lens distortion $d[K_1,K_2]$ for different blur levels. It also exhibits the effect of the approximate inversion of the distortion $u[K_1,K_2]$. Indeed, the inversion works quite well for most blur levels, except for some artifacts close to the image border. These become more pronounced for higher blur levels. This is due to fact that the forward and inverse distortion models \eqref{eq:distort} and \eqref{eq:undistort} presented in Section~\ref{sec:spatial_variance} are not exactly matching each other. In general, the distortion becomes more severe towards the image border, while additional artifacts are caused by boundary conditions.
However, as the text characters in the HDC dataset are always located in the center, these artifacts can be mostly neglected in the context of the challenge. Nevertheless, the visual warp remaining in the inversion for higher blur levels offers an explanation for the drop in performance of our reconstruction pipeline.

\subsection{Image Reconstruction}\label{subsec:results:reconstr}
\begin{figure}
	\centering
	\includegraphics[width=.8\linewidth]{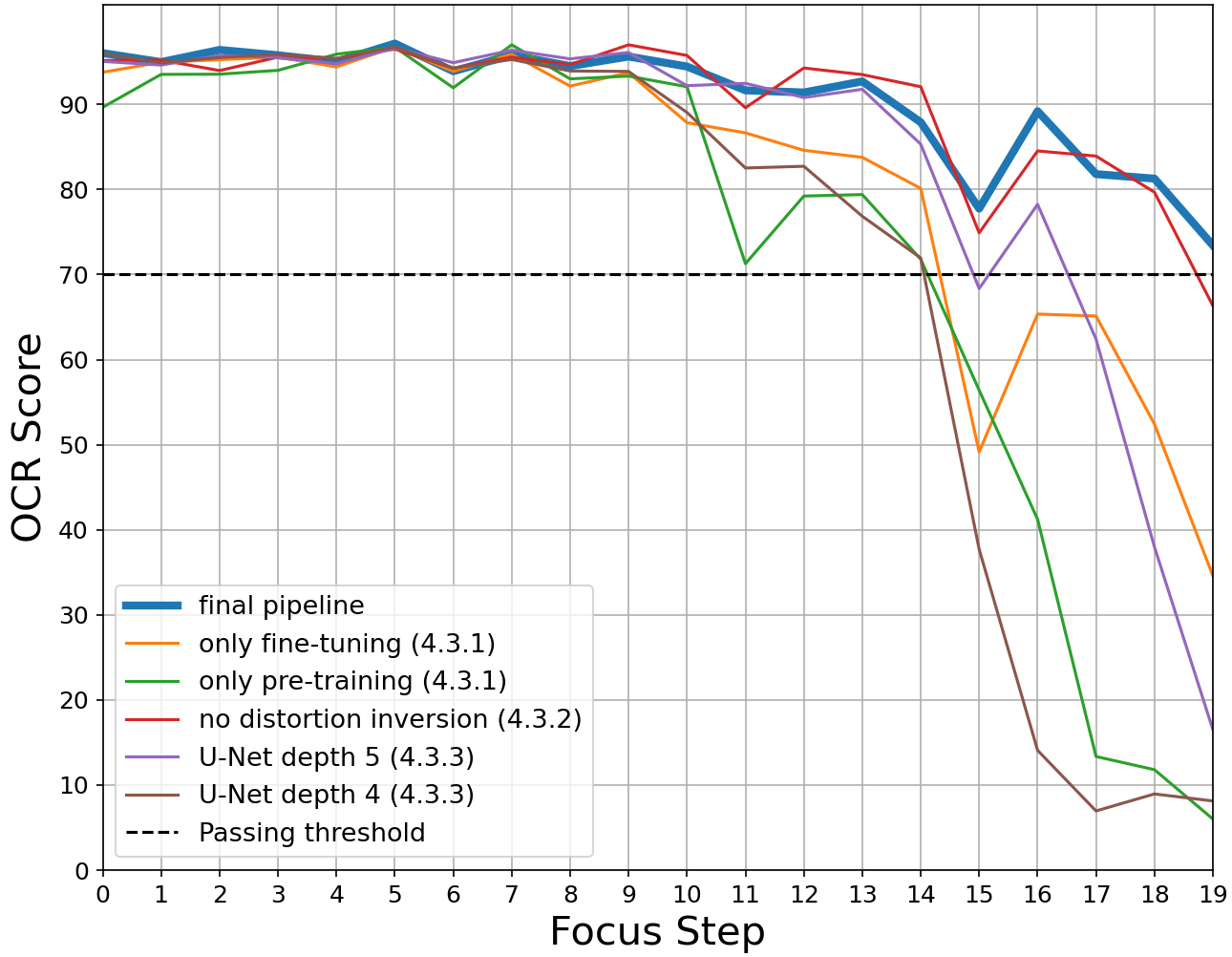}  
	\caption{Summary of OCR scores of all ablation studies presented in Section~\ref{subsec:results:reconstr}. Analogously to Fig.~\ref{fig:hdc_res}, the OCR scores are computed as the average over the challenge test set. Each line corresponds to a different modification of our final challenge submission, which is highlighted as the bold line. For analogous plots reporting SSIM and PSNR, see Fig.~\ref{fig:ssim_avg} and Fig.~\ref{fig:psnr_avg} in Appendix~\ref{sec:app:metrics}.}
	\label{fig:ablation_avg_ocr}
\end{figure}

This section presents three ablation studies analyzing the importance of several aspects of our reconstruction pipeline. We will focus on a visual assessment of the reconstruction results, but also quantitatively support our findings by reporting the associated OCR scores in Fig.~\ref{fig:ablation_avg_ocr}.\footnote{For technical reasons, we have slightly cropped the boundary of the reconstructed images before the OCR is performed. This explains the slight deviation of our evaluation results from the OCR scores computed by the challenge organizers in Fig. \ref{fig:hdc_res}.}

\subsubsection{Ablation Study: Pre-Training and Fine-Tuning}
\begin{figure} 
	\centering
	\begin{tabular}{c@{\,}c@{\,}c@{\,}c@{\,}c}
		
		\textbf{\ level \ } & blurry image & only pre-training & only fine-tuning & both \\
		\vspace{0.15em}
		
		\textbf{4} & 
		\includegraphics[valign=c,width=2.7cm]{img/fwd/naive/gt/gt_step_04.png} & 
		\includegraphics[valign=c,width=2.7cm]{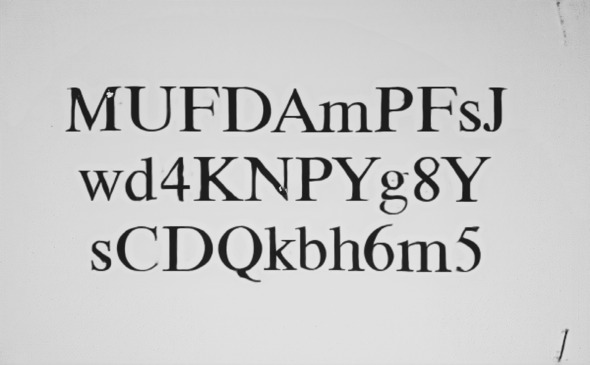} & 
		\includegraphics[valign=c,width=2.7cm]{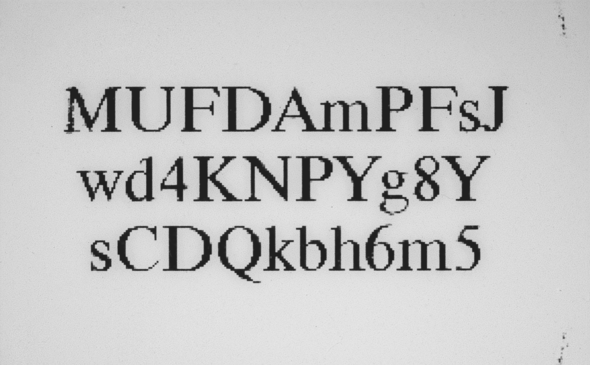} & 
		\includegraphics[valign=c,width=2.7cm]{img/rec/ours/FT/deblurred_FT=Truestep_04_OCR=None.png} \\      
		\vspace{0.15em}
		
		\textbf{9} & 
		\includegraphics[valign=c,width=2.7cm]{img/fwd/naive/gt/gt_step_09.png} & 
		\includegraphics[valign=c,width=2.7cm]{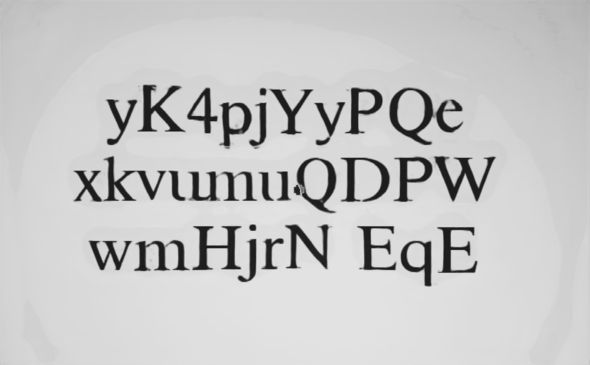} & 
		\includegraphics[valign=c,width=2.7cm]{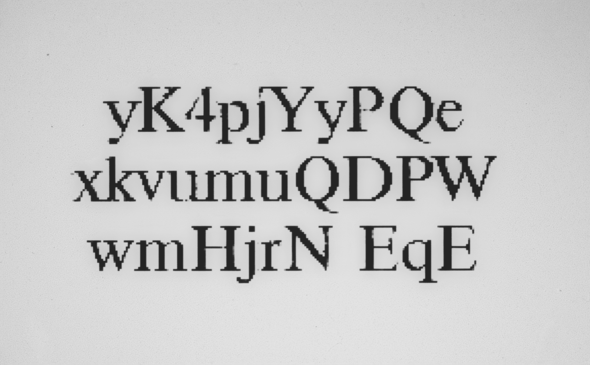} & 
		\includegraphics[valign=c,width=2.7cm]{img/rec/ours/FT/deblurred_FT=Truestep_09_OCR=100.0.png} \\
		\vspace{0.15em}
		
		\textbf{14} & 
		\includegraphics[valign=c,width=2.7cm]{img/fwd/naive/gt/gt_step_14.png} & 
		\includegraphics[valign=c,width=2.7cm]{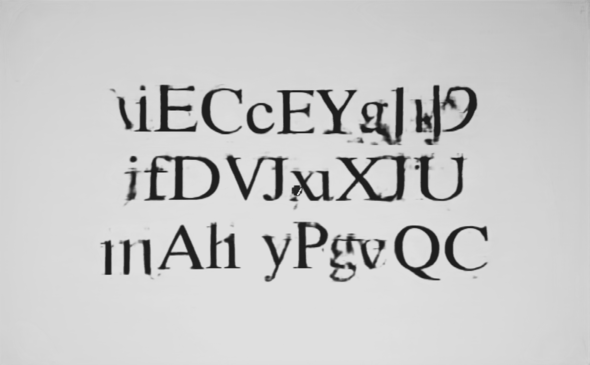} & 
		\includegraphics[valign=c,width=2.7cm]{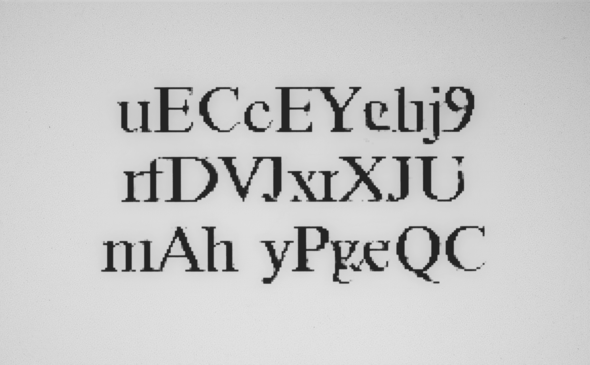} & 
		\includegraphics[valign=c,width=2.7cm]{img/rec/ours/FT/deblurred_FT=Truestep_14_OCR=76.0.png} \\
		\vspace{0.15em}
		
		\textbf{19} & 
		\includegraphics[valign=c,width=2.7cm]{img/fwd/naive/gt/gt_step_19.png} & 
		\includegraphics[valign=c,width=2.7cm]{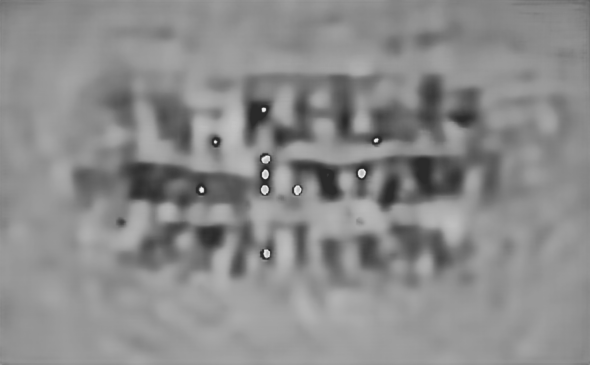} & 
		\includegraphics[valign=c,width=2.7cm]{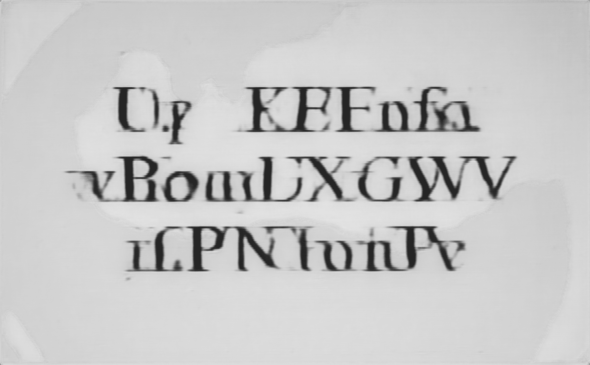} & 
		\includegraphics[valign=c,width=2.7cm]{img/rec/ours/FT/deblurred_FT=Truestep_19_OCR=67.0.png}
	\end{tabular}
	\caption{Comparison of deblurring networks trained only on synthetic data (= only pre-training), trained only on the original HDC data (= only fine-tuning), and trained as in Section~\ref{sec:step_4} (= both).}
	\label{fig:ablation_pretrain}
\end{figure}

The first study relates to our training scheme, comparing the performance of our final challenge submission to identical network architectures that were trained only on the synthetic pre-training data and only on the original HDC data, respectively. We evaluate all networks on random image samples from the HDC test set.
Fig.~\ref{fig:ablation_pretrain} provides a qualitative comparison of the results. Our main findings are as follows:
\begin{enumerate}
	\item 
	\emph{Only pre-training (second column):} Even though the original HDC data is never observed in this case, it is remarkable that respectable reconstructions are achieved for most blur levels. The network only fails to produce reasonable results for the last few levels. We suspect that the generalization capacity from synthetic data is closely linked to the quality of the estimated forward model.
	Indeed, our experiments from Section~\ref{subsec:results:fwd} reveal that the forward model estimation is quite accurate for the lower levels while there is a noticeable drop in quality in the higher ones.
	\item 
	\emph{Only fine-tuning (third column):} The network only trained on the provided HDC data leads to a higher reconstruction quality compared to the pre-training-only variant. However, it still fails in the last few levels. Although severe overfitting is to be expected from training only on $178$ image pairs, it is remarkable that the network still manages to deblur digits, which were never observed during training, e.g., see level $14$ in Fig.~\ref{fig:ablation_pretrain}.
	\item
	\emph{Both (right column):}
	The training scheme used for our challenge submission (pre-training on synthetic data and fine-tuning on mostly HDC data as described in Section~\ref{sec:step_4}) provides a significant improvement in reconstruction quality, which becomes particularly visible from blur level $9$ on.
\end{enumerate}
Given the effectiveness of our pre-training/fine-tuning approach (see Section~\ref{sec:step_4}), the following ablation studies are all based on this training scheme.

\subsubsection{Ablation Study: Inverse Distortion}

\begin{figure}
	\centering
	\begin{tabular}{c@{\,}c@{\,}c@{\,}c}
		\textbf{\ level \ } & blurry image & no inversion & with inversion \\
		\vspace{0.15em}
		
		\textbf{4} & 
		\includegraphics[valign=c,width=2.7cm]{img/fwd/naive/gt/gt_step_04.png} & 
		\includegraphics[valign=c,width=2.7cm]{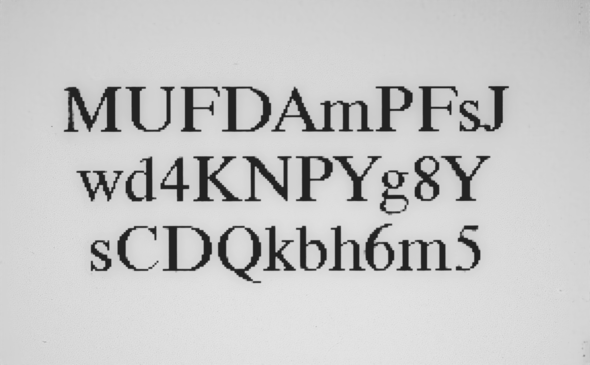} & 
		\includegraphics[valign=c,width=2.7cm]{img/rec/ours/FT/deblurred_FT=Truestep_04_OCR=None.png}  \\
		\vspace{0.15em}
		
		\textbf{9} & 
		\includegraphics[valign=c,width=2.7cm]{img/fwd/naive/gt/gt_step_09.png} & 
		\includegraphics[valign=c,width=2.7cm]{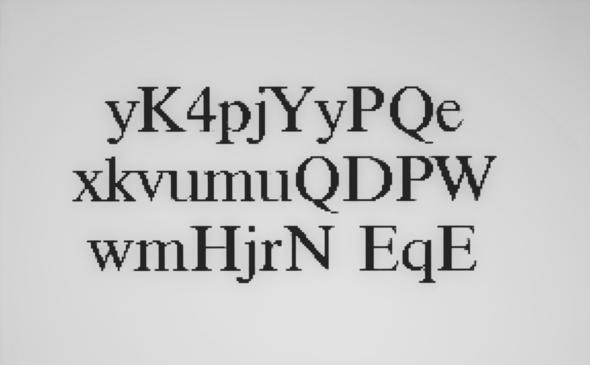} & 
		\includegraphics[valign=c,width=2.7cm]{img/rec/ours/FT/deblurred_FT=Truestep_09_OCR=100.0.png} \\
		\vspace{0.15em}
		
		\textbf{14} & 
		\includegraphics[valign=c,width=2.7cm]{img/fwd/naive/gt/gt_step_14.png} & 
		\includegraphics[valign=c,width=2.7cm]{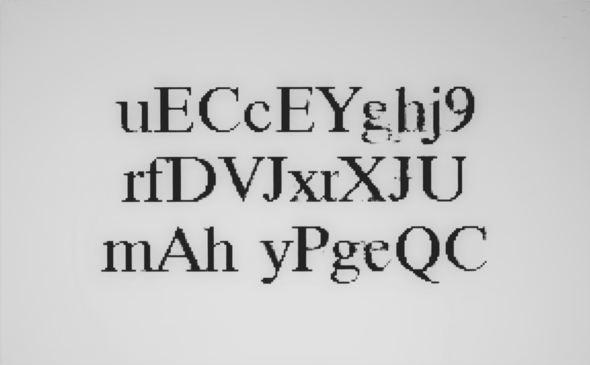} & 
		\includegraphics[valign=c,width=2.7cm]{img/rec/ours/FT/deblurred_FT=Truestep_14_OCR=76.0.png} \\
		\vspace{0.15em}
		
		\textbf{19} & 
		\includegraphics[valign=c,width=2.7cm]{img/fwd/naive/gt/gt_step_19.png}& 
		\includegraphics[valign=c,width=2.7cm]{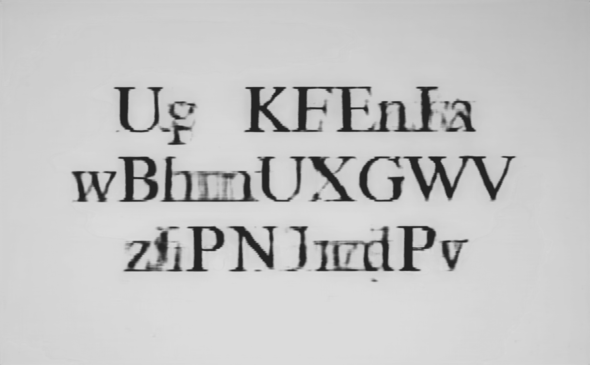} & 
		\includegraphics[valign=c,width=2.7cm]{img/rec/ours/FT/deblurred_FT=Truestep_19_OCR=67.0.png}
	\end{tabular}
	\caption{Comparison of deblurring U-Nets without initial inverse radial distortion (middle column) and with initial inverse radial distortion as described in Section~\ref{sec:step_3} (right column); see Fig.~\ref{fig:ablation_undist_zoom} for a zoom of the results of level 19.}
	\label{fig:ablation_undist}
\end{figure}

\begin{figure}
	\centering
	\includegraphics[width=.8\linewidth]{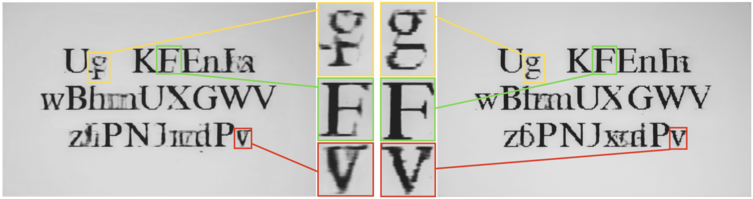}  
	\caption{Zoom of level $19$ from Fig.~\ref{fig:ablation_undist}, comparing the deblurring results of U-Nets without (left) and with (right) inverse radial distortion as an initial step.}
	\label{fig:ablation_undist_zoom}
\end{figure}

In order to investigate the effect of pre-processing the blurry input images with approximately inverting the radial distortion (see Section~\ref{sec:step_3}), we have trained the deblurring U-Net with and without this step; see Fig.~\ref{fig:ablation_undist} and Fig.~\ref{fig:ablation_undist_zoom} for results. 
Compared to the previous ablation study, differences in reconstruction quality are less significant, but we observe that inverting the radial lens distortion consistently produces sharper reconstructions.
For example, see the sharper letters in blur level~$19$, which seem to smear a little without inverted distortion. This slight improvement is also supported quantitatively, as we can observe in Fig.~\ref{fig:ablation_avg_ocr}, since we would not have cleared the last challenge level without the inversion step.

\subsubsection{Ablation Study: U-Net Depth}

\begin{figure} 
	\centering
	\begin{tabular}{@{}c@{\,}c@{\,}c@{\,}c@{\,}c@{}} 
		\textbf{\ level \ } & blurry image & depth = $4$ & depth = $5$ & depth = $6$ \\
		\vspace{0.15em}
		
		\textbf{9} & 
		\includegraphics[valign=c,width=2.7cm]{img/fwd/naive/gt/gt_step_09.png} &
		\includegraphics[valign=c,width=2.7cm]{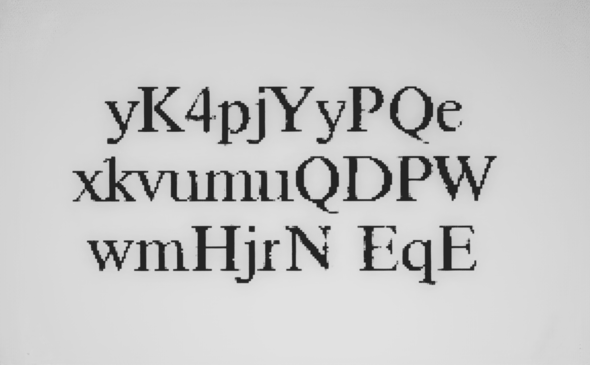} & 
		\includegraphics[valign=c,width=2.7cm]{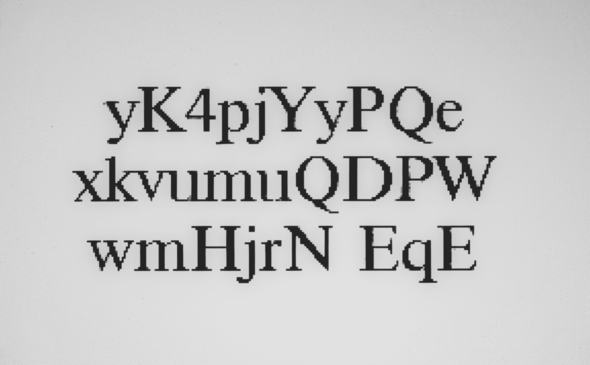} & 
		\includegraphics[valign=c,width=2.7cm]{img/rec/ours/FT/deblurred_FT=Truestep_09_OCR=100.0.png}  \\
		\vspace{0.15em}
		
		\textbf{14} & 
		\includegraphics[valign=c,width=2.7cm]{img/fwd/naive/gt/gt_step_14.png} &
		\includegraphics[valign=c,width=2.7cm]{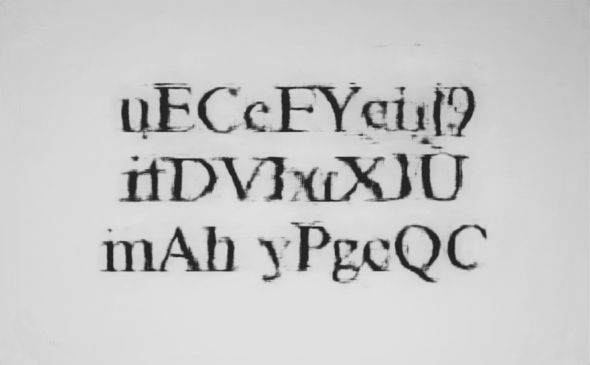} & 
		\includegraphics[valign=c,width=2.7cm]{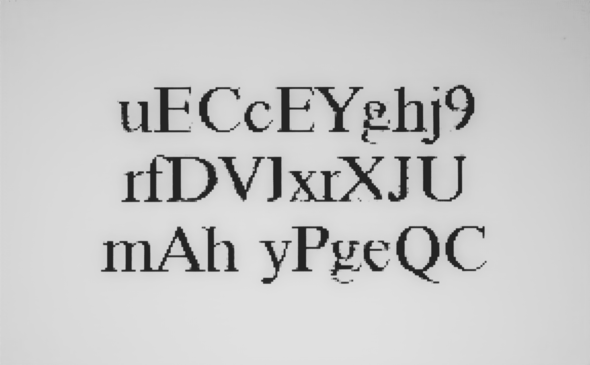} & 
		\includegraphics[valign=c,width=2.7cm]{img/rec/ours/FT/deblurred_FT=Truestep_14_OCR=76.0.png} \\
		\vspace{0.15em}
		
		\textbf{17} & 
		\includegraphics[valign=c,width=2.7cm]{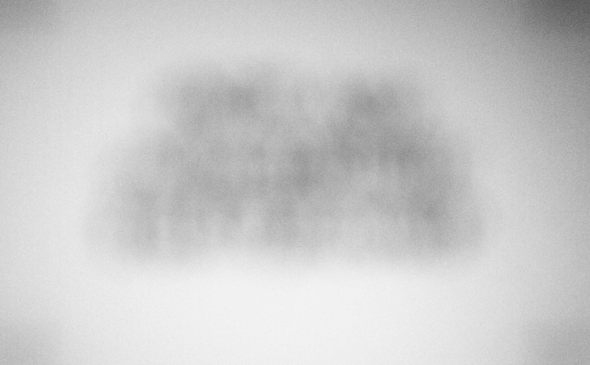} &
		\includegraphics[valign=c,width=2.7cm]{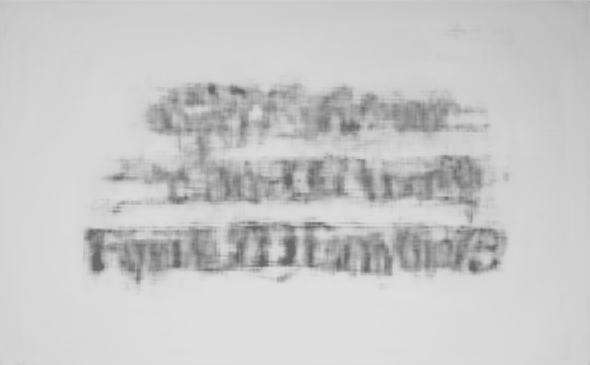} & 
		\includegraphics[valign=c,width=2.7cm]{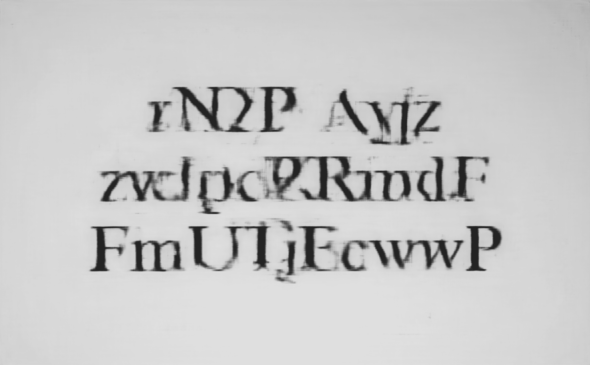} & 
		\includegraphics[valign=c,width=2.7cm]{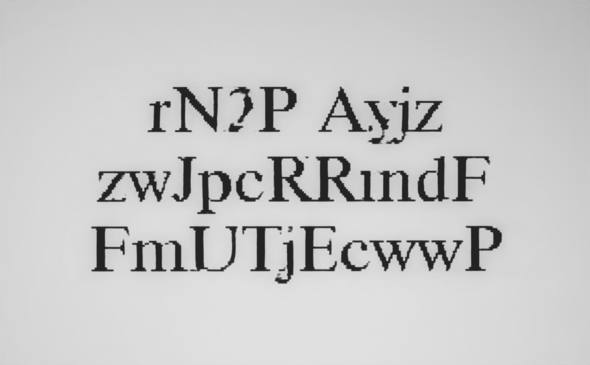} \\
		\vspace{0.15em}
		
		\textbf{19} & 
		\includegraphics[valign=c,width=2.7cm]{img/fwd/naive/gt/gt_step_19.png} &
		\includegraphics[valign=c,width=2.7cm]{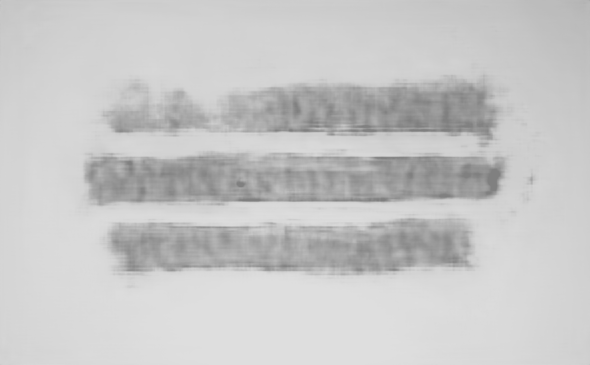} & 
		\includegraphics[valign=c,width=2.7cm]{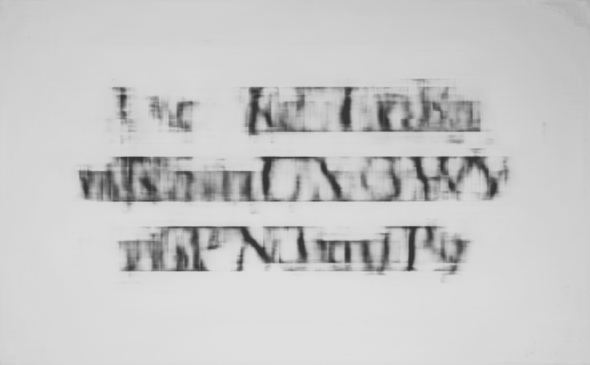} & 
		\includegraphics[valign=c,width=2.7cm]{img/rec/ours/FT/deblurred_FT=Truestep_19_OCR=67.0.png}
	\end{tabular}
	\caption{Comparison of deblurring U-Nets of different depths (= number of down- and up-sampling steps).}
	\label{fig:ablation_depth}
\end{figure}

We now study how the ``depth'' of the U-Net affects the deblurring performance. To this end, we train three versions of the U-Net (all with inverted distortion) differing only in the number of down- and up-sampling steps, which we call the \emph{U-Net depth} --- more specifically, the depths $4$, $5$, and $6$ are considered. 
Recall that the vanilla U-Net~\cite{U-Net} has a depth of $4$ and our submitted U-Net has depth $6$ (see also Section~\ref{sec:step_3}). Due to memory constraints, we were not able to train even deeper U-Nets. 
Fig.~\ref{fig:ablation_depth} presents our experimental results.
In contrast to the other experiments, the blur levels $9$, $14$, $17$, and $19$ are displayed here, as they visualize more clearly how depth improves the deblurring performance. 
This observation underpins the need for a large receptive field to capture the size of increasingly wider blur kernels.

\section{Conclusion and Discussion}\label{sec:disc}

We have developed a competitive, state-of-the-art deblurring algorithm that is capable of processing severe degrees of defocus blur.
A key point of our solution to the HDC is a solid understanding of modeling and estimating the forward operator.
This allowed us to build a high-quality stream of synthetic data, which greatly benefited the generalization capacity of the reconstruction network as well as our architectural design choices. 
Those include the inverted radial distortion to render the problem approximately translation invariant and the deeper U-Net to enlarge the receptive field.
In the following, we discuss some open problems and potential directions for future research.

\subsection{Technical Aspects}
On a technical note, it would be interesting to see how well deeper U-Nets would perform, especially on an even stronger defocus blur. Another direction to study is increasing the kernel size of the U-Net convolutions, which is fixed to the standard choice $3$ in our case. Both approaches help to enlarge the receptive field and it seems natural to explore their interplay and trade-offs, also in the context of other imaging tasks. 
More specific to our deblurring problem is the design of more elaborate distortion models and related inversion mechanisms, which holds the promise of achieving even sharper reconstructions on the last few blur levels.

\subsection{The Bigger Picture}

While the positive impact of synthetic data is well-known in deep learning research \cite{Nikolenko:2019aa}, we have demonstrated it in the context of solving inverse problems.
However, our focus is on the specific task of image deblurring, and therefore, further research is needed to draw more general conclusions on the benefits of synthetic data.

Furthermore, the stability of our algorithmic pipeline remains to be investigated. As our methodology was deployed in a highly controlled challenge environment, its practical applicability to other types of blurred images is not guaranteed --- the HDC sanity check was just a first step in that direction.
One important concern is an extension of our approach to different camera setups, i.e., different forward operators.
This could be realized by an appropriate calibration step, which affects the weights of the blur kernel and radial lens distortion. 
Some preliminary experiments on the robustness of our pipeline can be found in Appendix~\ref{sec:app:robust}.
Another interesting avenue of future research would be to develop a ``one-size-fits-all'' reconstruction network that can handle arbitrary degrees of blurring. 
One conceivable way to do so would be to combine the outputs of all $20$ trained U-Nets and processing them in a subsequent CNN.
Ideas from knowledge distillation could be also useful in that regard \cite{Hinton:2015aa}.

To move beyond laboratory conditions towards a production-ready model, many other aspects need to be taken into account. Some examples are varying levels of luminance within the images, covariate shifts, motion blur, tangential distortion effects due to misaligned cameras, and more diverse data to build a general-purpose deblurrer.

Having said that, the purpose of this article is \emph{not} to pave the way to an algorithm that can be directly implemented into a product. 
Instead, it should be rather seen as an important proof of concept for future undertakings, closely aligned with the original goals of the HDC. 
We have demonstrated the potential of data-centric machine learning and physics-informed modeling in inverse problems, thereby pushing the limits beyond what was previously considered achievable --- admittedly, the authors, and perhaps the challenge organizers as well, initially did not expect that passing a blur level far higher than 10 would be possible.\footnote{In the run-up to the official start of the challenge, the organizers only collected data for the first 10 blur levels, which quickly turned out to be too ``easy''. This motivated them to significantly increase the severity of blurring by adding 10 more levels, up to fairly extreme situations.}
We hope that our findings can spark theoretical and empirical research that goes beyond classical/sparse regularization theory (cf.~Section~\ref{sec:literature}) and puts the role of data even more in the foreground than it was already done in compressed sensing.

\section*{Acknowledgments}
We would like to thank Samuli Siltanen, Markus Juvonen, Fernando Moura as well as the Finish Inverse Problems Society for organizing this beautiful and inspiring challenge.
Moreover, we would like to thank Gabriele Steidl for bringing the team together and supporting us in this research project.
Finally, we would like to thank the anonymous reviewers for their comments and suggestions, which have helped to improve the presentation of this paper.

\appendix

\section{Additional Experiments}

\subsection{Standard Imaging Metrics}
\label{sec:app:metrics}

\begin{figure}[H]
	\centering
	\includegraphics[width=.7\linewidth]{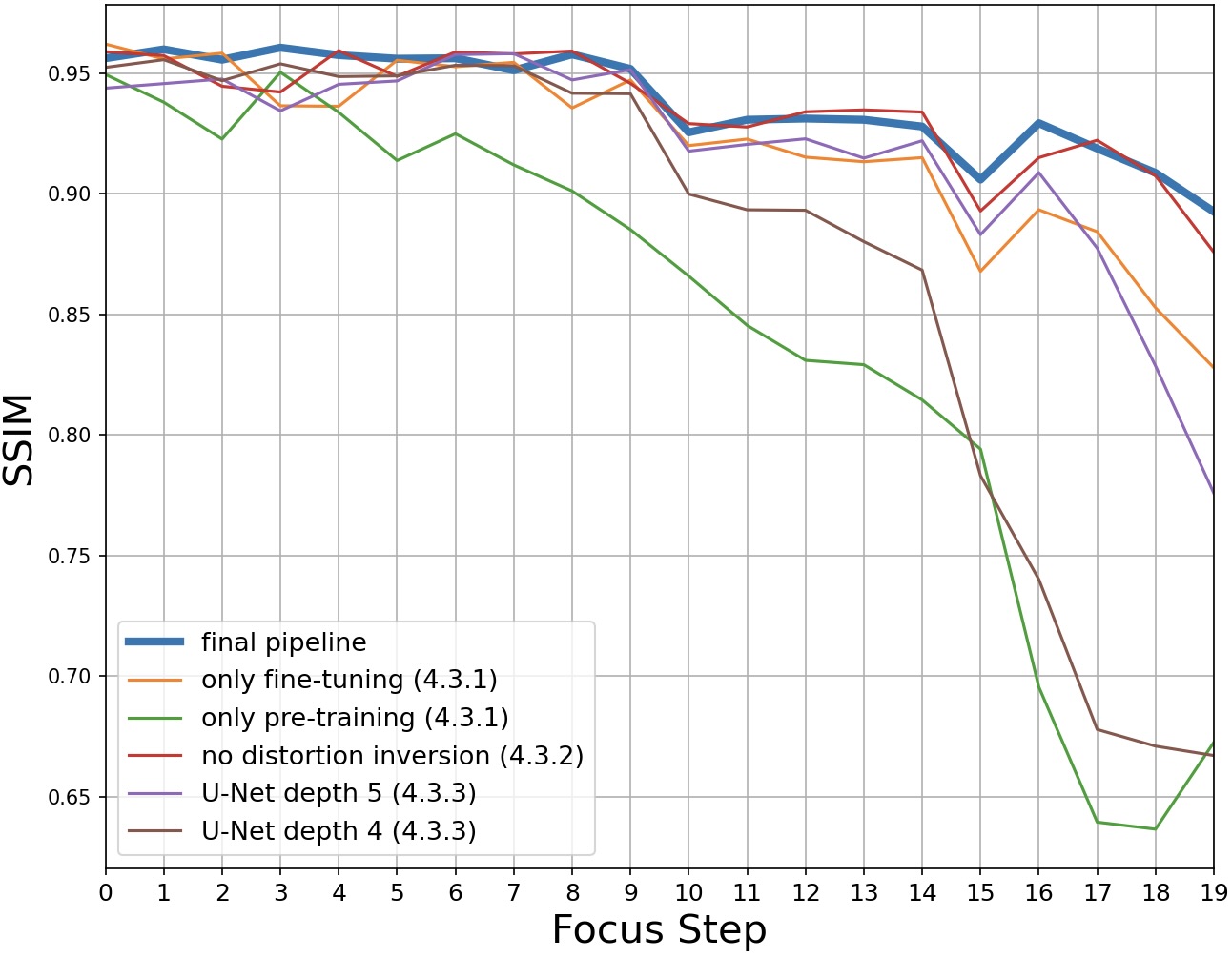} 
	\caption{Average SSIM scores for our ablation studies from Section~\ref{subsec:results:reconstr}; cf.~Fig.~\ref{fig:ablation_avg_ocr}.}
	\label{fig:ssim_avg}
\end{figure}

\begin{figure}[H]
	\centering
	\includegraphics[width=.7\linewidth]{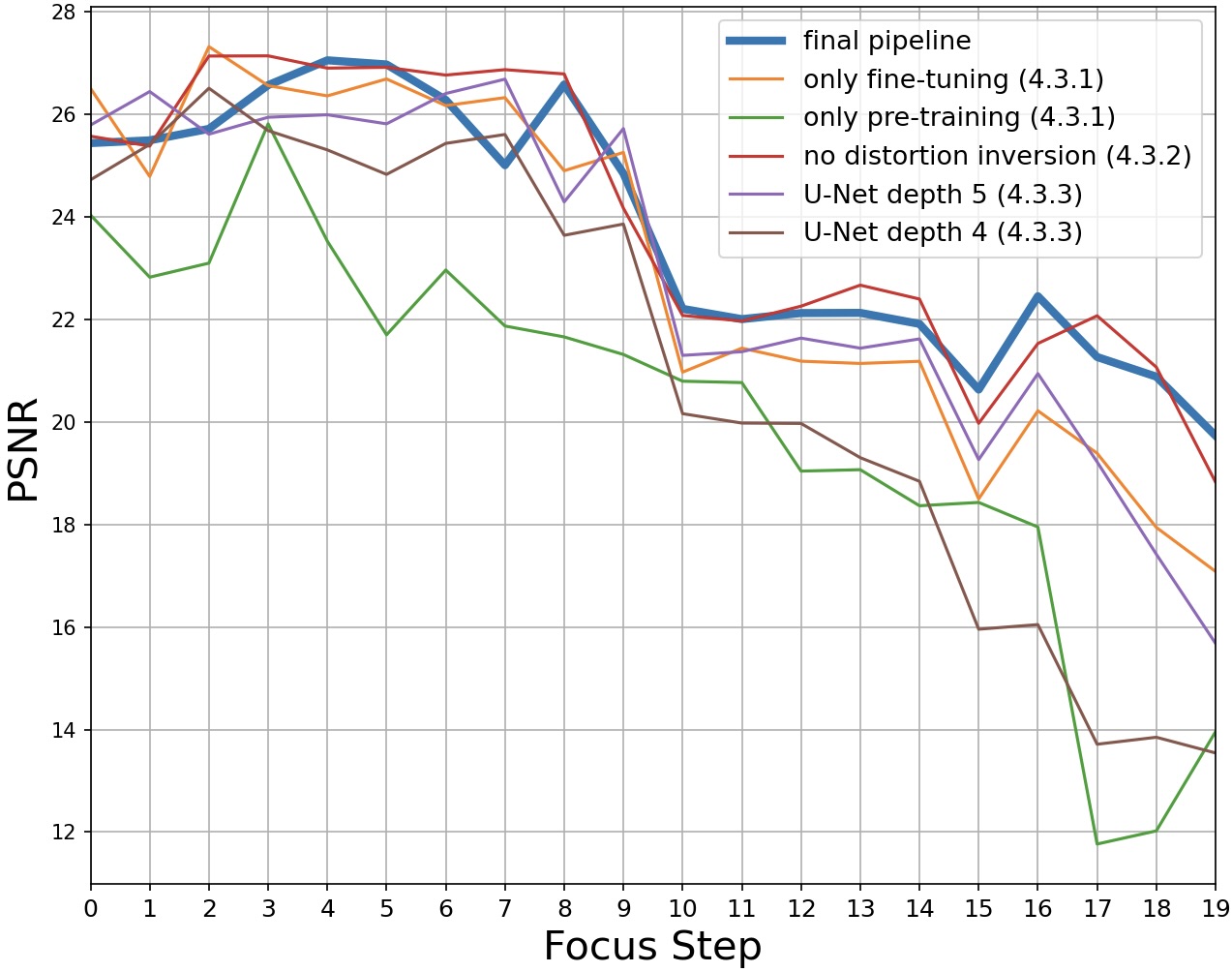} 
	\caption{Average PSNR scores for our ablation studies from Section~\ref{subsec:results:reconstr}; cf.~Fig.~\ref{fig:ablation_avg_ocr}.}
	\label{fig:psnr_avg}
\end{figure}

This section supplements our ablation study from Section~\ref{subsec:results:reconstr}. To allow for better comparability with existing and future approaches, we report the average structural similarity index measure (SSIM) in Fig.~\ref{fig:ssim_avg} and the average peak signal-to-noise ratio (PSNR) in Fig.~\ref{fig:psnr_avg}, both computed on the test dataset.

\subsection{Out-Of-Distribution Robustness}
\label{sec:app:robust}

\begin{figure}
	\centering
	\begin{tabular}{c@{\,}c@{\,}c@{\,}c@{\,}c}
		\textbf{\ level \ } & blurry image & $i-2$ reconstr.\ & $i-1$ reconstr.\ & reconstruction \\
		\vspace{0.15em}
		
		\textbf{4} & 
		
		\includegraphics[valign=c,width=2.7cm]{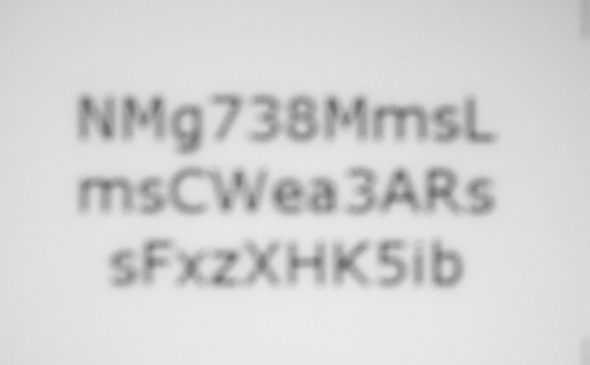} &  
		\includegraphics[valign=c,width=2.7cm]{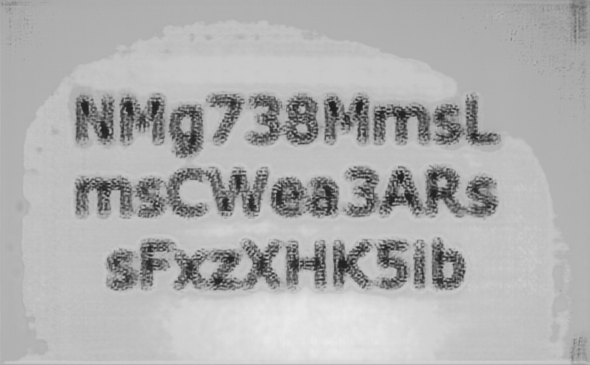} & 
		\includegraphics[valign=c,width=2.7cm]{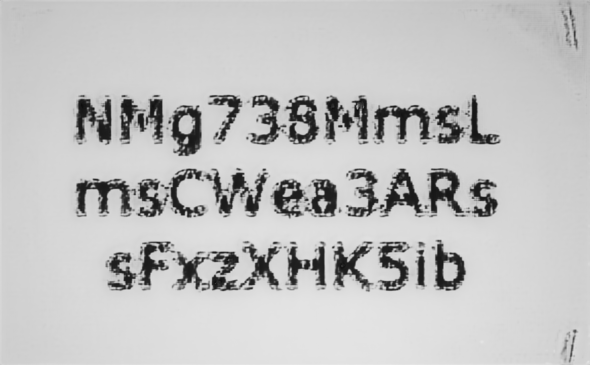} & 
		\includegraphics[valign=c,width=2.7cm]{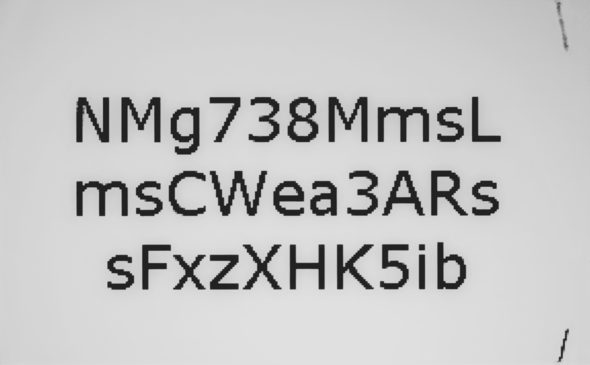} \\
		
		\vspace{0.15em}
		
		\textbf{9} & 
		
		\includegraphics[valign=c,width=2.7cm]{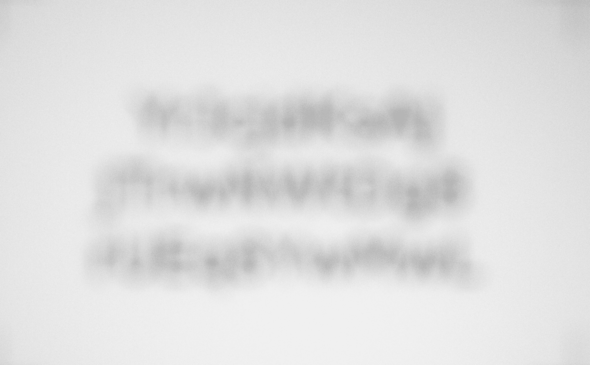} & 
		\includegraphics[valign=c,width=2.7cm]{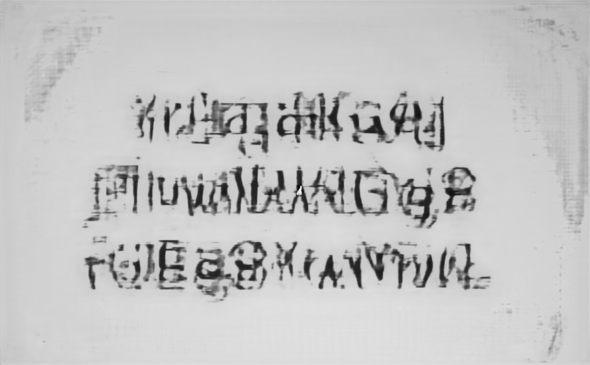} & 
		\includegraphics[valign=c,width=2.7cm]{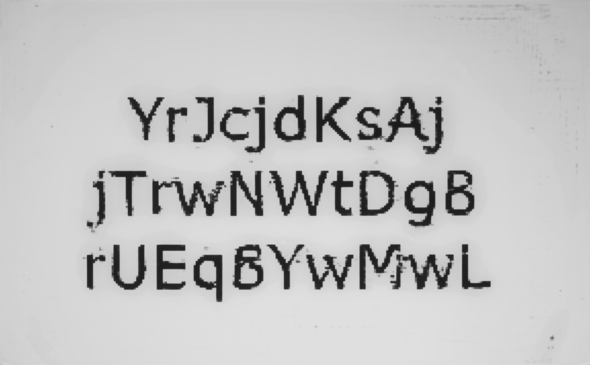} & 
		\includegraphics[valign=c,width=2.7cm]{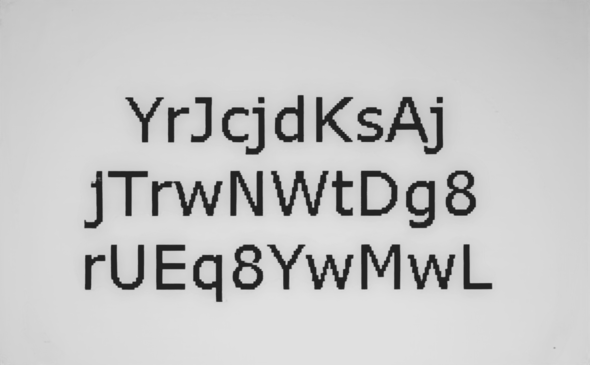}
		\\
		\vspace{0.15em}
		
		\textbf{14} & 
		
		\includegraphics[valign=c,width=2.7cm]{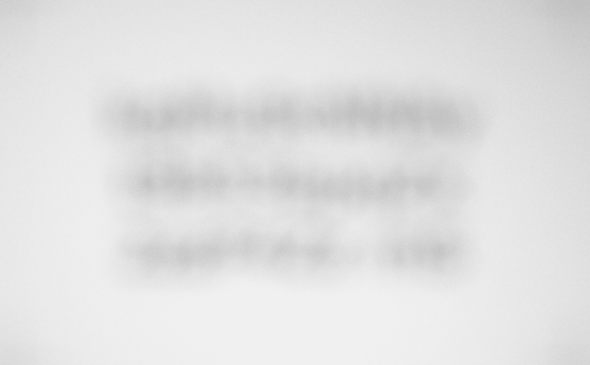} & 
		\includegraphics[valign=c,width=2.7cm]{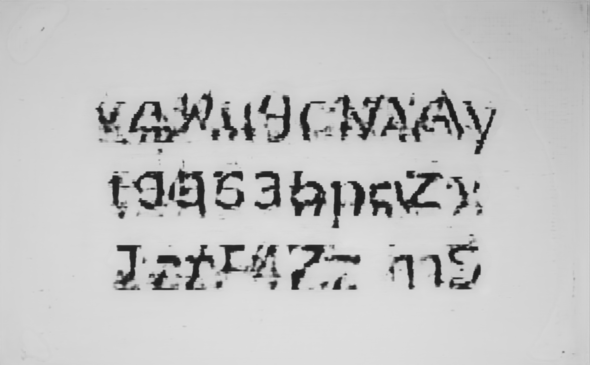} & 
		\includegraphics[valign=c,width=2.7cm]{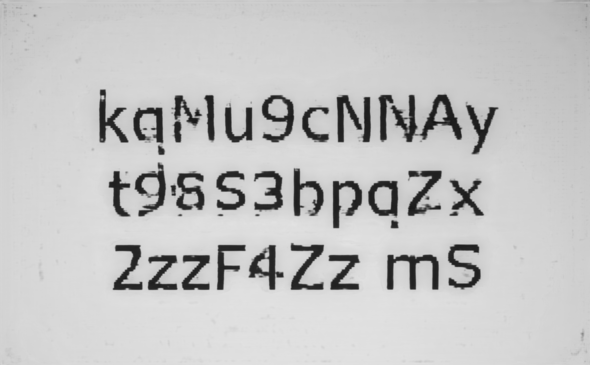} & 
		\includegraphics[valign=c,width=2.7cm]{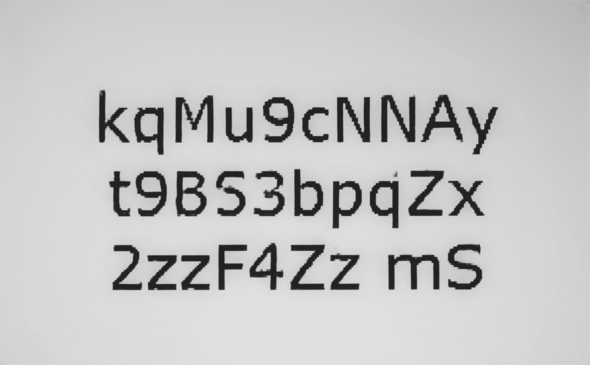}
		\\
		\vspace{0.15em}
		
		\textbf{17} &   
		\includegraphics[valign=c,width=2.7cm]{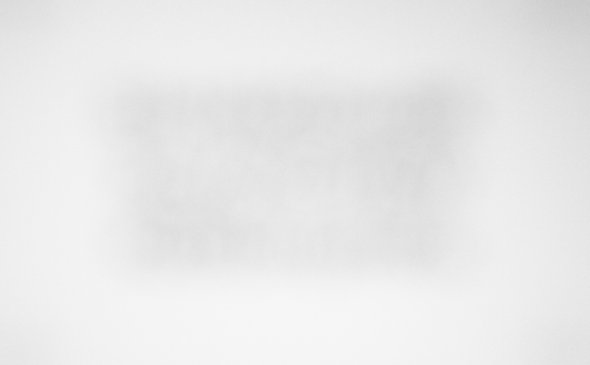} & 
		\includegraphics[valign=c,width=2.7cm]{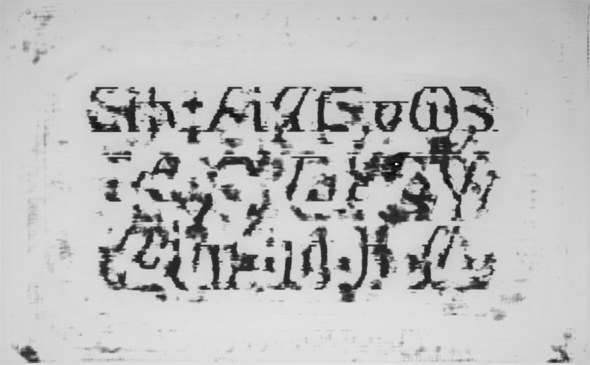} & 
		\includegraphics[valign=c,width=2.7cm]{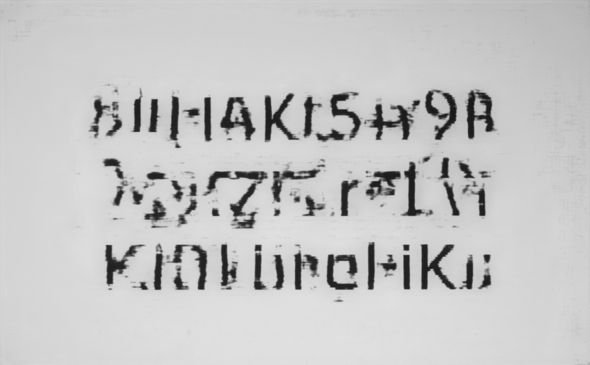} & 
		\includegraphics[valign=c,width=2.7cm]{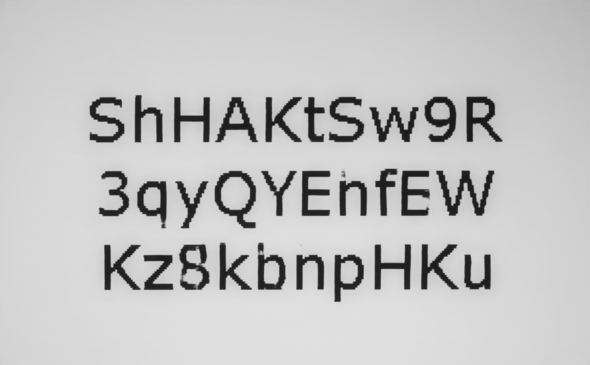} 
		\\
		\vspace{0.15em}

		\textbf{19} & 
		\includegraphics[valign=c,width=2.7cm]{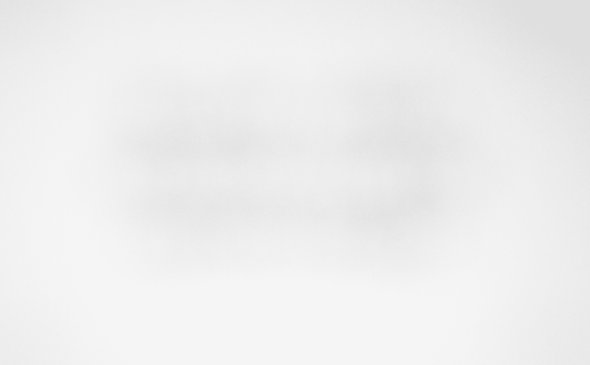} & 
		\includegraphics[valign=c,width=2.7cm]{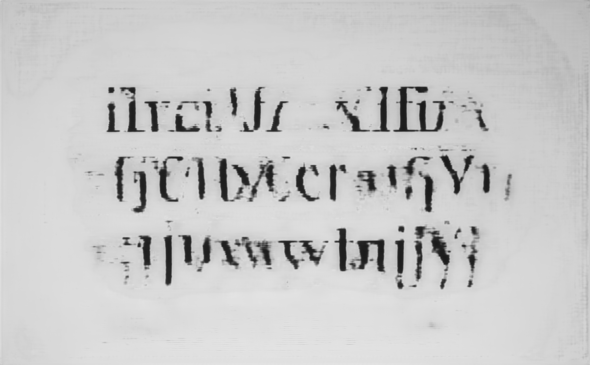} & 
		\includegraphics[valign=c,width=2.7cm]{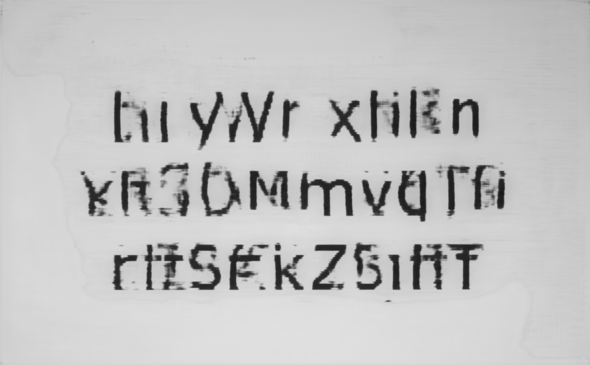} & 
		\includegraphics[valign=c,width=2.7cm]{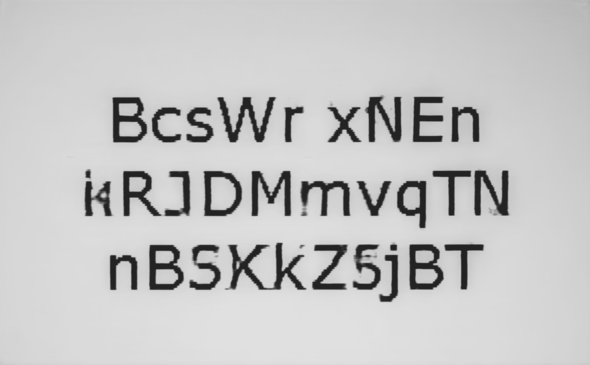} 
		\\
		\vspace{0.15em}

	\end{tabular}
	\caption{Out-of-distribution performance on examples from the test dataset. The images are taken from the blur levels $i = 4$, $9$, $14$, $17$, and $19$, while the used reconstruction pipelines have been trained on the easier levels $i-1$ and $i-2$ (right column = pipeline of level $i$ for reference).}
	\label{fig:ood_minus}
\end{figure}

\begin{figure}
	\centering
	\begin{tabular}{c@{\,}c@{\,}c@{\,}c@{\,}c}
		\textbf{\ level \ } & blurry image & $i+1$ reconstr.\ & $i+2$ reconstr.\ & reconstruction \\
		\vspace{0.15em}
		
		\textbf{4} & 
		
		\includegraphics[valign=c,width=2.7cm]{img/review/blurry_step_04.png} &  
		\includegraphics[valign=c,width=2.7cm]{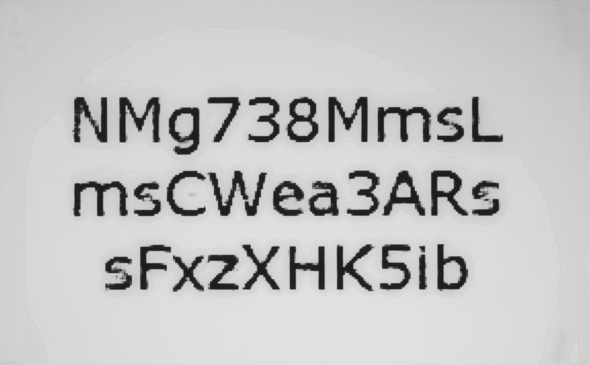} & 
		\includegraphics[valign=c,width=2.7cm]{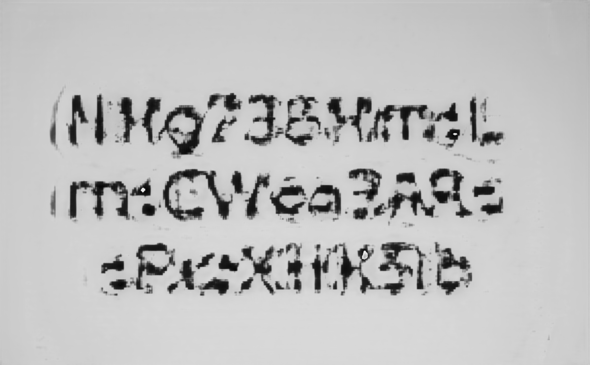} & 
		\includegraphics[valign=c,width=2.7cm]{img/review/deblurred_FT=Truestep_04.png} \\
		
		\vspace{0.15em}
		
		\textbf{9} & 
		
		\includegraphics[valign=c,width=2.7cm]{img/review/blurry_step_09.png} & 
		\includegraphics[valign=c,width=2.7cm]{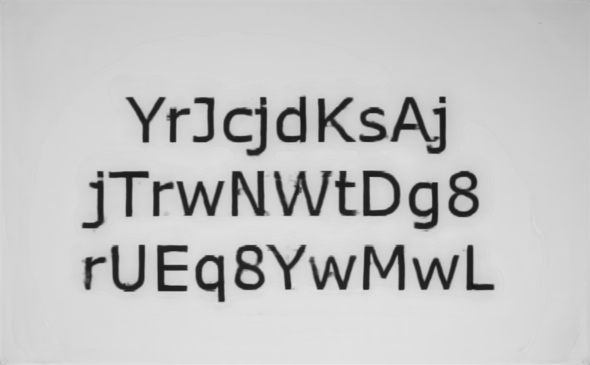} & 
		\includegraphics[valign=c,width=2.7cm]{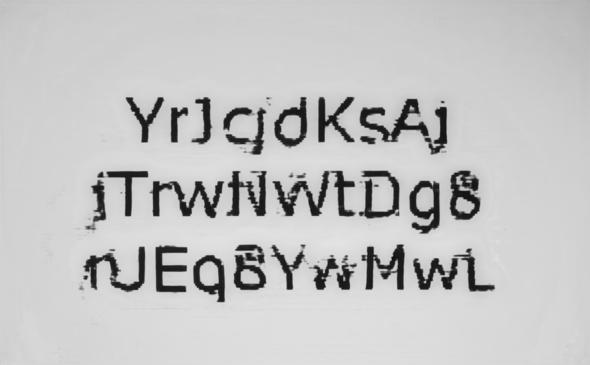} & 
		\includegraphics[valign=c,width=2.7cm]{img/review/deblurred_FT=Truestep_09.png}
		\\
		\vspace{0.15em}
		
		\textbf{14} & 
		
		\includegraphics[valign=c,width=2.7cm]{img/review/blurry_step_14.png} & 
		\includegraphics[valign=c,width=2.7cm]{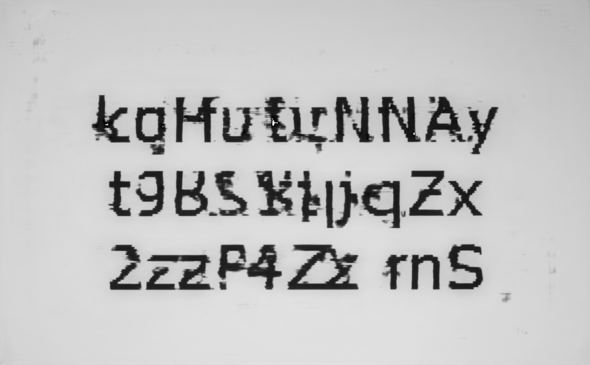} & 
		\includegraphics[valign=c,width=2.7cm]{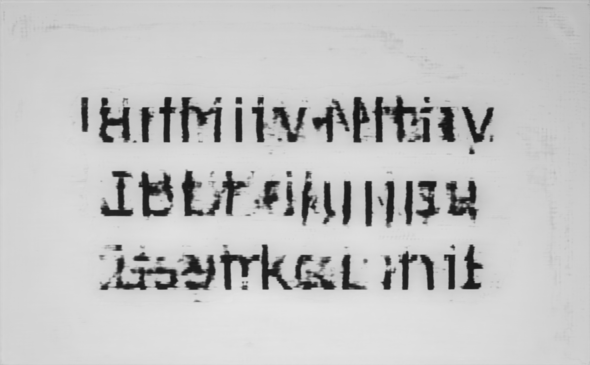} & 
		\includegraphics[valign=c,width=2.7cm]{img/review/deblurred_FT=Truestep_14.png}
		\\
		\vspace{0.15em}

		\textbf{17} & 
		
		\includegraphics[valign=c,width=2.7cm]{img/review/blurry_step_17.png} & 
		\includegraphics[valign=c,width=2.7cm]{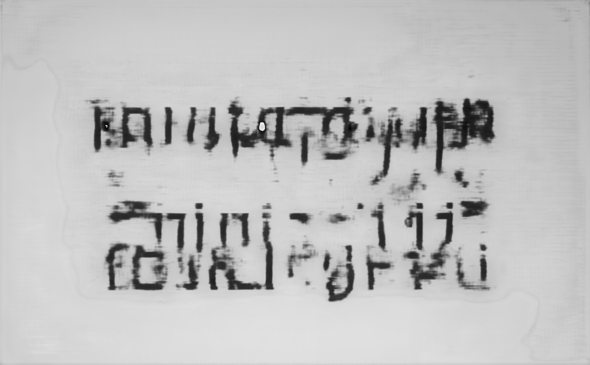} & 
		\includegraphics[valign=c,width=2.7cm]{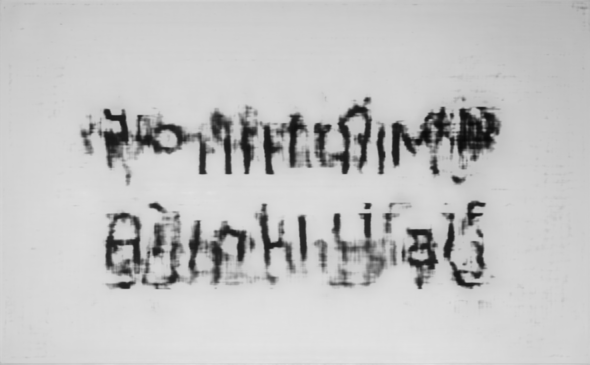} & 
		\includegraphics[valign=c,width=2.7cm]{img/review/deblurred_FT=Truestep_17.png} 
		\\
		\vspace{0.15em}

	\end{tabular}
	\caption{Out-of-distribution performance on examples from the test dataset. The images are taken from the blur levels $i = 4$, $9$, $14$, and $17$, while the used reconstruction pipelines have been trained on the harder levels $i+1$ and $i+2$ (right column = pipeline of level $i$ for reference). Note that for level $i = 19$, there exists no corresponding pipelines, which explains why we have included level $17$ instead.}
	\label{fig:ood_plus}
\end{figure}

\begin{figure}
	\centering
	\includegraphics[width=.8\linewidth]{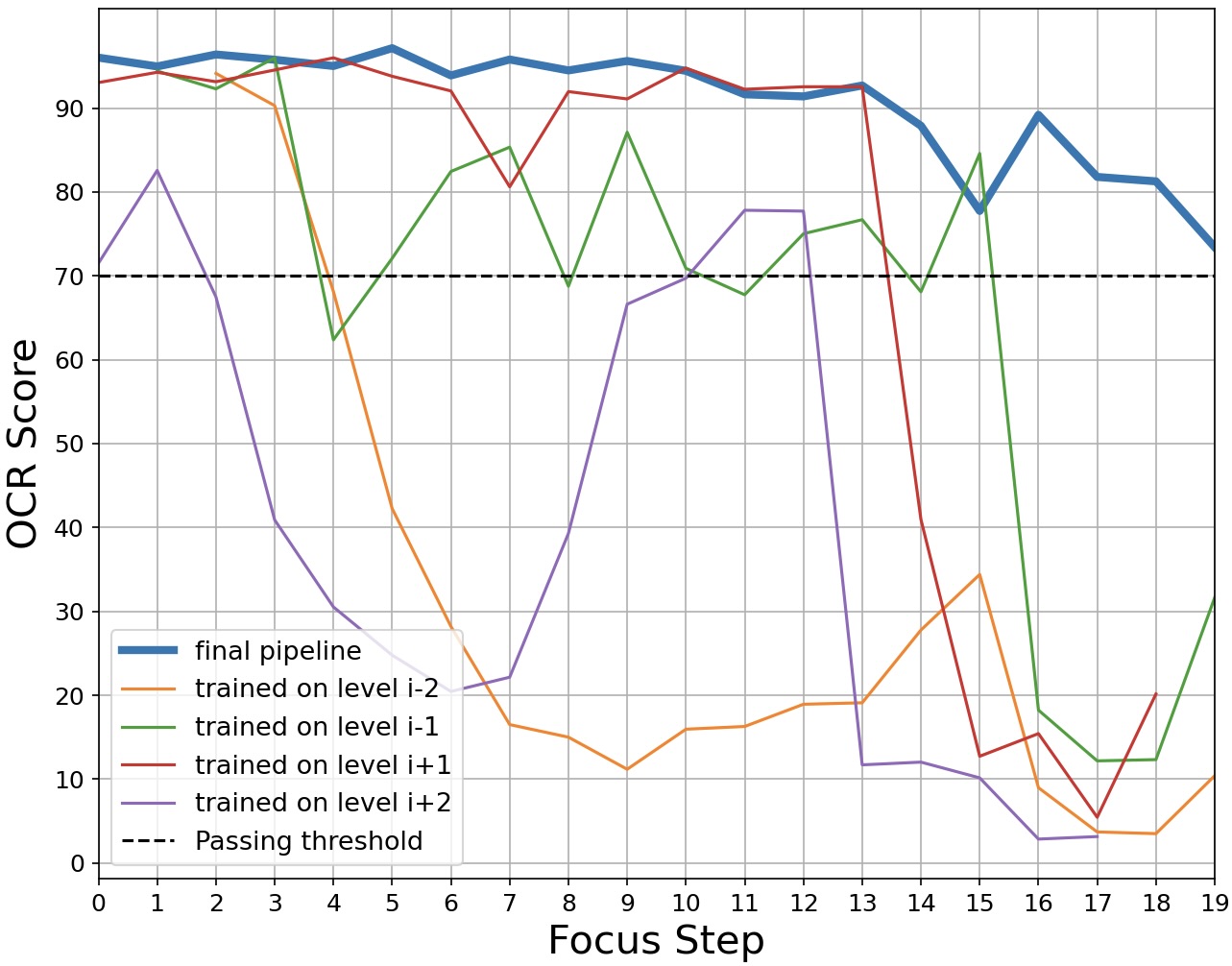} 
	\caption{Average OCR Scores on the test data for our final pipeline but applied to adjacent levels.}
	\label{fig:OOD}
\end{figure}

In an effort to assess the robustness of our solution, we have taken test images from a given level $i$ and reconstructed them with the pipelines of levels $i-2$, $i-1$, $i+1$, and $i+2$, respectively. Fig.~\ref{fig:ood_minus} considers a fixed image from level $i$ and shows the corresponding reconstructions for the pipelines trained on the easier deblurring tasks of level $i-1$ and $i-2$. The analogous experiment for the harder deblurring tasks of level $i+1$ and $i+2$ is shown in Fig.~\ref{fig:ood_plus}. In Fig.~\ref{fig:OOD}, these results are quantified by reporting the average OCR scores across all (possible) levels.
In a nutshell, we conclude that the out-of-distribution performance is decent for the lower blur levels and also when reconstructing with the pipeline from a directly adjacent blur level.

\clearpage

\bibliographystyle{abbrv}
\bibliography{reference.bib}

\medskip

\end{document}